\pdfoutput=1
\documentclass{article}
\usepackage{arxiv}

\usepackage{verbatim}
\usepackage{amssymb}
\usepackage{pifont}
\usepackage{xcolor}
\usepackage{multirow}
\usepackage{array}
\usepackage{calc}
\usepackage{makecell}
\usepackage{relsize}
\usepackage{microtype}
\usepackage{amsthm}
\usepackage{graphicx}
\usepackage{float}
\usepackage{booktabs}
\usepackage{mathrsfs}  
\usepackage{multirow}
\usepackage{amsmath}
\usepackage{multirow}

\usepackage{placeins}
\usepackage{tabularx}
\usepackage{subcaption}
\usepackage{graphicx}
\usepackage[normalem]{ulem}

\useunder{\uline}{\ul}{}

\usepackage[utf8]{inputenc} 
\usepackage[T1]{fontenc}    
\usepackage{hyperref}       
\usepackage{url}            
\usepackage{booktabs}       
\usepackage{amsfonts}       
\usepackage{nicefrac}       
\usepackage{microtype}      
\usepackage{lipsum}		
\usepackage[square,sort,comma,numbers]{natbib}
\usepackage{doi}
\usepackage{xcolor}
\usepackage{multirow}
\usepackage{array}
\usepackage{calc}
\usepackage{makecell}
\usepackage{relsize}
\usepackage{microtype}
\usepackage{amsthm}
\usepackage{graphicx}
\usepackage{float}
\usepackage{booktabs}
\usepackage{mathrsfs}  
\usepackage{multirow}
\usepackage{amsmath}
\usepackage{multirow}
\usepackage{placeins}
\usepackage{tabularx}
\usepackage{subcaption}
\usepackage{graphicx}
\usepackage[normalem]{ulem}

\usepackage{verbatim}
\usepackage{amssymb}
\usepackage{pifont}
\usepackage{xcolor}
\usepackage{multirow}
\usepackage{array}
\usepackage{calc}
\usepackage{makecell}
\usepackage{relsize}
\usepackage{microtype}
\usepackage{amsthm}
\usepackage{graphicx}
\usepackage{float}
\usepackage{booktabs}
\usepackage{mathrsfs}  
\usepackage{multirow}
\usepackage{amsmath}
\usepackage{multirow}
\usepackage{placeins}

\usepackage{tabularx}
\usepackage{subcaption}
\usepackage{graphicx}
\usepackage[normalem]{ulem}

\useunder{\uline}{\ul}{}

\useunder{\uline}{\ul}{}

\title{Domain Generalization - A Causal Perspective}

\author{Paras Sheth, Raha Moraffah, K. Sel\c{c}uk Candan, Adrienne Raglin$^{\dagger}$, Huan Liu
\thanks{Adrienne Raglin is with Army Research Laboratory (ARL). All other authors are with School of Computing and Augmented Intelligence, Arizona State University, Tempe, AZ, USA. (e-mail: \{psheth5,rmoraffa,candan,huanliu\}@asu.edu, adrienne.raglin2.civ@mail.mil)}}

\renewcommand{\shorttitle}{\textit{arXiv} Template}

\newtheorem{definition}{Definition}

\makeatletter

\hypersetup{
pdftitle={A template for the arxiv style},
pdfsubject={q-bio.NC, q-bio.QM},
pdfauthor={David S.~Hippocampus, Elias D.~Striatum},
pdfkeywords={First keyword, Second keyword, More},
}

\begin{document}
\maketitle

\begin{abstract}
	Machine learning models rely on various assumptions to attain high accuracy. One of the preliminary assumptions of these models is the independent and identical distribution, which suggests that the train and test data are sampled from the same distribution. However, this assumption seldom holds in the real world due to distribution shifts. As a result models that rely on this assumption exhibit poor generalization capabilities. Over the recent years, dedicated efforts have been made to improve the generalization capabilities of these models collectively known as -- \textit{domain generalization methods}. The primary idea behind these methods is to identify stable features or mechanisms that remain invariant across the different distributions. Many generalization approaches employ causal theories to describe invariance since causality and invariance are inextricably intertwined. However, current surveys deal with the causality-aware domain generalization methods on a very high-level. Furthermore, we argue that it is possible to categorize the methods based on how causality is leveraged in that method and in which part of the model pipeline is it used. To this end, we categorize the causal domain generalization methods into three categories, namely, (i) Invariance via Causal Data Augmentation methods which are applied during the data pre-processing stage, (ii) Invariance via Causal representation learning methods that are utilized during the representation learning stage, and (iii) Invariance via Transferring Causal mechanisms methods that are applied during the classification stage of the pipeline. Furthermore, this survey includes in-depth insights into benchmark datasets and code repositories for domain generalization methods. We conclude the survey with insights and discussions on future directions.
\end{abstract}


Machine learning (ML) models have achieved widespread success in variety of applications, including recommender systems~\cite{sheth2022causal}, autonomous cars~\cite{stilgoe2018machine}, and across various areas including, but not limited to, computer vision, graphs, and natural language processing. However, the success of these models is accompanied by various assumptions such as the independent and identical distributions or the i.i.d. assumption. According to this assumption the training and testing data are identically and independently distributed. In other words, the train and test data stem from the same distribution.

\begin{figure}[t!]
    \centering
    \includegraphics[width=0.95\linewidth]{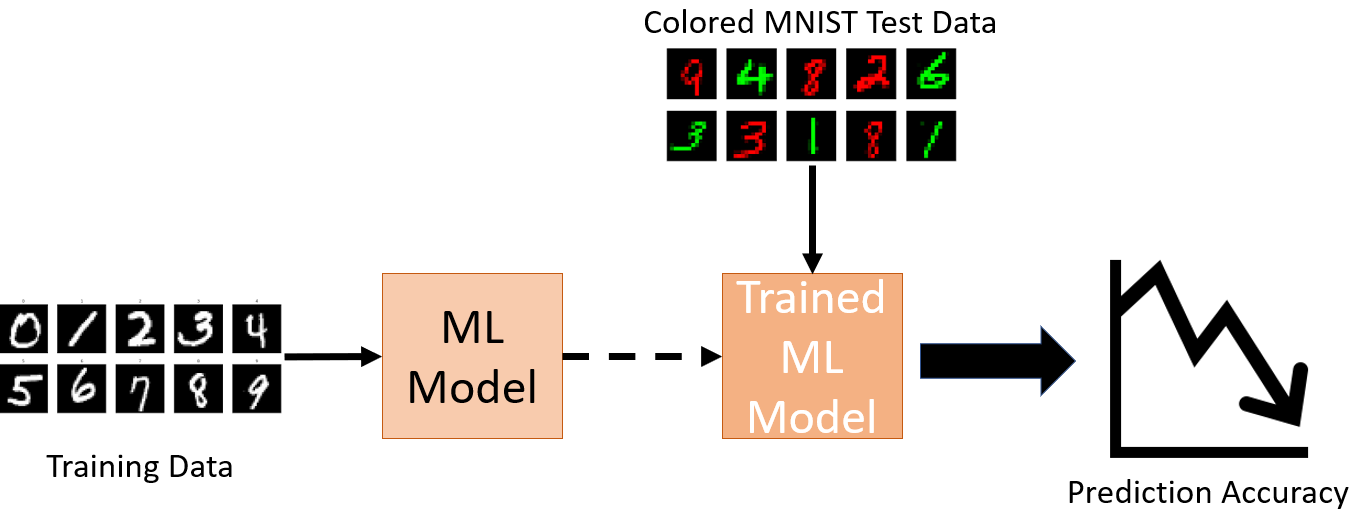}
    \caption{The task of digit classification from a domain generalization perspective. A model trained on MNIST dataset would fail to generalize when evaluated on ColoredMNIST, or RotatedMNIST.}
    \label{fig1}
\end{figure}

However, real-world data seldom abides by this assumption.  Due to the dynamic nature of the systems that employ the machine learning models, the training data distribution might not be the same as the test data distribution and as a result the model’s accuracy decreases. Under the machine learning paradigm, this phenomenon is known as \textit{distribution shift}. For instance consider the task of digit classification as shown in Fig.~\ref{fig1}. Given a model trained on a set of black and white handwritten digits~\cite{lecun1998gradient} and evaluated on a set of colored handwritten digits~\cite{arjovsky2019invariant} it is observed that the model performance drastically reduces, which raises the question \textit{Why this happens?} It is well known that machine learning models leverage correlations among the different features to improve the prediction accuracy~\cite{pettit2021artificial,demler2013impact}. However, in multiple real-world scenarios and the shown example, the model learns patterns (correlations) that are present in the training data but do not hold when evaluated against the test data. Thus, it results in the degrading performance.

One possible solution to address the distribution shift problem is to improve the \textit{generalization} capabilities of the ML models which can be done in various ways including Domain Adaptation (DA) and Domain Generalization (DG). Although domain adaptation methods lie outside the scope of this survey, we introduce them to distinguish between domain generalization and domain adaptation. Generally, both DA and DG focus on the same task, i.e., to improve the generalizability in ML models. However, the key difference between DA and DG methods is that the DA methods use a small number of unlabeled data points from the previously unseen test data (also known as target domain(s)) to fine-tune the model trained on the training data (also known as source domain(s))~\cite{saenko2010adapting,lu2020stochastic,liu2020open}. However, in real-world settings the target data is usually not accessible beforehand to perform model adaptation. Thus, to deal with situations when the target domain data is unavailable, \textit{domain generalization} methods were introduced~\cite{blanchard2011generalizing}.

With the distinction of DG and DA methods, we focus on understanding the working mechanism of DG methods, and how causality aids in the process. To understand this, let’s consider the aforementioned digit classification scenario. Although the train domain data consisted of black and white images of digits and the target domain consisted of colored digits, as humans we are still able to classify the digits even after the distribution shift. 
Even though the digits are colored in the target domain, the shape of the digit, which is crucial in determining the digit, remains invariant. This implies that even in the presence of distribution shifts there exists a set of features that are invariant and are crucial for the prediction performance. Furthermore, it is well established that causality and invariance are tightly linked to each other, i.e., one of the dimensions of causality is invariance~\cite{bourrat2021measuring,buhlmann2020invariance}. Thus,  causality can be a useful tool in capturing the invariance present in the data, justifying the range of methods that have leveraged different causal theories for improving the generalization capabilities of models~\cite{mahajan2021domain,arjovsky2019invariant}.

\begin{figure}
    \centering
    \includegraphics[width=\linewidth]{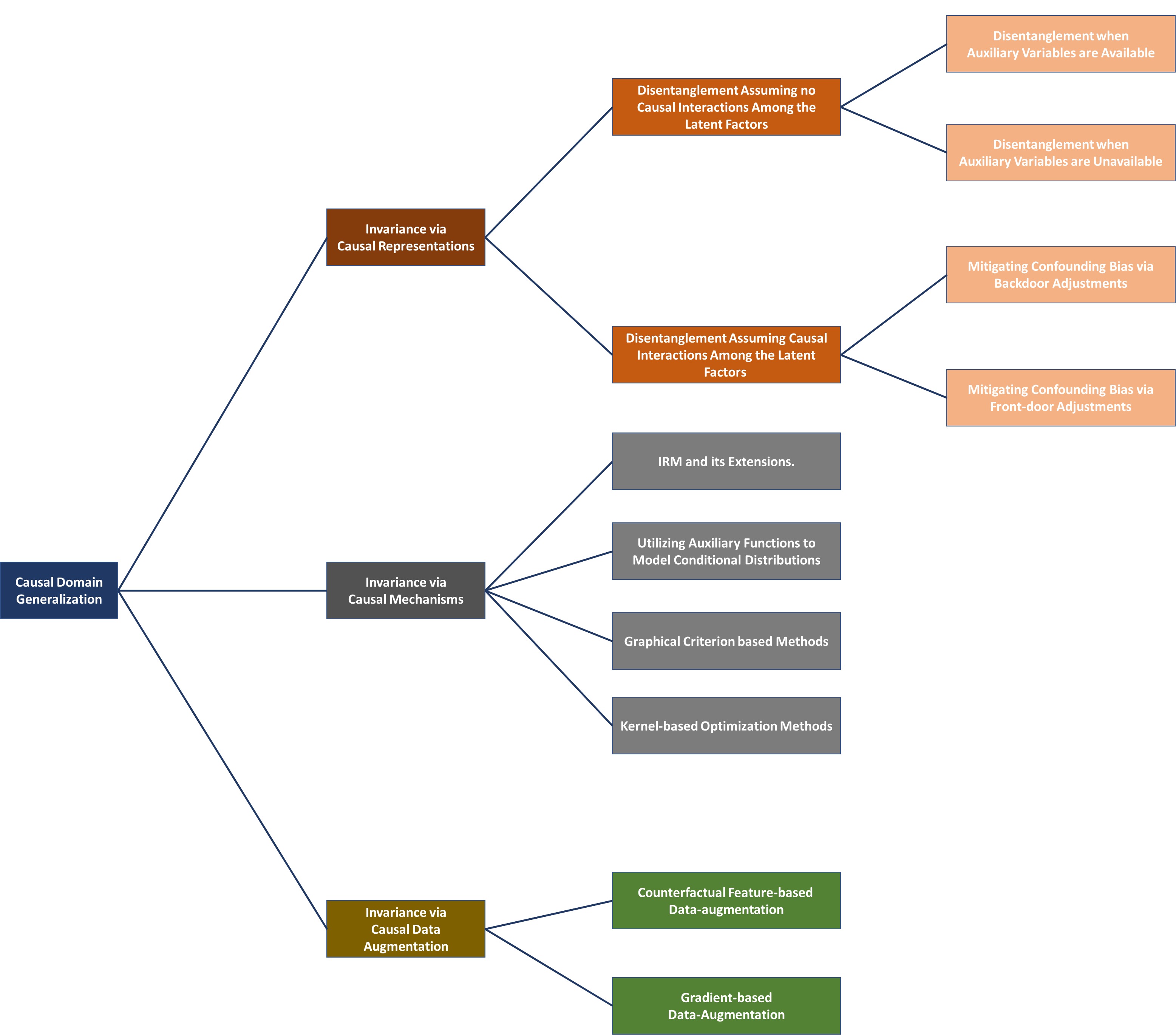}
    \caption{The categorization of Causal Domain Generalization techniques.} 
    \label{categorization}
\end{figure}

Since there exist multiple causality aware domain generalization methods, the next step should be to understand how these methods leverage causality and how they differ among themselves. Although majority of the existing surveys discuss causal domain generalization methods~\cite{zhou2022domain,shen2021towards,wang2022generalizing} they mostly club all the methods under the same umbrella term, i.e., causality aware domain generalization methods or causal representation methods. We argue that these methods can be better classified based on where the causal theories are leveraged during the entire model pipeline. To this end, we categorize these papers broadly into three different categories: (i) Invariance via Causal Data Augmentation methods which are applied during the data pre-processing stage, (ii) Invariance via Causal representation learning methods that are utilized during the representation learning stage, and (iii) Invariance via Transferring Causal mechanisms methods that are applied during the classification stage of the pipeline. Furthermore, Fig.~\ref{categorization} shows the detailed breakdown of these categories into their corresponding subcategories.

The remaining part of the survey is categorized as follows. Section~\ref{pdp} covers the problem definition and causal preliminaries, Section~\ref{cat} covers the different categories and the sub-categories helping the reader understand how causality can be leveraged in different parts of the model pipeline and how each category tackles its corresponding challenges. Section~\ref{cat_tg} covers the Causal DG methods developed for the graphs and Natural Language Processing (NLP). Section~\ref{bench} covers the benchmark datasets and evaluation schemes employed by various methods for evaluating the performances. This section also covers the publicly available code repositories which could aid researchers in developing and testing their own models. We finally conclude the survey with Section~\ref{conc}.

\section{Problem Definition and Preliminaries}
\label{pdp}
\subsection{Problem Definition}
Consider $X$ as the set of features, $Y$ as the set of labels, and $D$ as the set of domain(s) with sample spaces $\mathcal{X}$, $\mathcal{Y}$, and $\mathcal{D}$, respectively. A domain is defined as a joint distribution $P_{X, Y}$ on $\mathcal{X} \times \mathcal{Y}$. Let $P_{X}$ represent $X$'s marginal distribution, $P_{X \mid Y}$ represent the class-conditional distribution of $X$ given $Y$, and  $P_{Y \mid X}$ represent the posterior distribution of $Y$ given $X$. 

A domain generalization model's purpose is to learn a predictive model $f: \mathcal{X} \rightarrow \mathcal{Y}$. However, while dealing with domain generalization, the common assumption implies that training data was obtained from a finite subset of the possible domains $D_{train} \subset \mathcal{D}$. Furthermore, the number of training domains is given by $K$, and $D_{train}=\{d_{i}\}_{i=1}^{K} \subset \mathcal{D}$. As a result, the training data is sampled from a distribution $P[X, Y \mid D=d_{i}] \quad \forall i \in [k]$. The domain generalization model, then aims at utilizing only source (train) domain(s) data with the goal of minimizing the prediction error on a previously unseen target (test) domain. The corresponding joint distribution of the target domain $D_{test}$ is given by $P_{X, Y}^{D_{test}}$ and $P_{X, Y}^{D_{test}} \neq P_{X Y}^{(k)}, \forall k \in\{1, \ldots, K\}$. Ideally, the goal is to learn a classifier that is optimal for all domains $\mathcal{D}$.
\subsection{Preliminaries}
Most causality-aware domain generalization models aim to mitigate the reliance on spurious correlations by accounting for the confounding effects. For instance, some works mitigate the confounding bias induced by the non-causal features on the causal features and the labels. This leads to better generalizability of the machine learning model. Confounding variables refer to those sets of variables that may be observed or unobserved, directly influencing the supposed cause and effect variables.

The confounding bias can be accounted for with either backdoor adjustment or front-door adjustment based on the nature of the problem. Therefore, we begin by defining the backdoor criterion.
\begin{definition}
 Given a Directed Acyclic Graph (DAG) $G$, a set of nodes $Z$, and a pair of nodes, namely, $X$ and $Y$, we say that $Z$ satisfies the backdoor criterion relative to $X$ and $Y$ if:
\begin{itemize}
\item no node in $Z$ is a descendant of $Y$
\item Every path between $X$ and $Y$ that contains an arrow in $Y$ is blocked by $Z$.
\end{itemize}
\end{definition}
If $Z$ satisfies the backdoor criterion for $X$ and $Y$, the causal effect between $X$ and $Y$ is identifiable. The causal effect from $X$ to $Y$ can be formulated as,
\begin{equation}
P(Y \mid X)=\sum_{z} P(Y \mid X, Z) P(Z).
\label{backdoor_real}
\end{equation}

Similarly, another way to account for the confounding variables is the front-door criterion. 
\begin{definition}
 Given a Directed Acyclic Graph (DAG) $G$, a set of nodes $Z$, and a pair of nodes, namely, $X$ and $Y$, we say that $Z$ satisfies the front-door criterion relative to $X$ and $Y$ if:
\begin{itemize}
\item All directed paths from $X$ to $Y$ are intercepted by Z;
\item There exists no unblocked backdoor path from $X$ to $Z$; and
\item $ X$ blocks all possible backdoor paths from $Z$ to $Y$.
\end{itemize}
\end{definition}
As per the front-door adjustment, if $Z$ satisfies the front-door criterion for $X$ and $Y$, and if $P(X,Z) > 0$, then the causal effect between $X$ and $Y$ is identifiable. The causal effect from $X$ to $Y$ can be formulated as,
\begin{equation}
P(Y \mid \operatorname{do}(X))=\sum_{z} \sum_{x^{\prime}} P(Y \mid Z, X^{\prime}) P(X^{\prime}) P(Z \mid X).
\label{frontdoor}
\end{equation}
\section{Domain Generalization via Invariance Learning}
\label{cat}
Causal Domain Generalization methods aim to leverage causal theories to improve the generalization capabilities of models. The causal theories are utilized in various stages of the standard machine learning pipeline. To this end, we categorize the Causal DG methods based on how and where the causality aspects are utilized. To this end, we propose three categories, namely, (1) Invariance via Causal Data Augmentation methods which are applied during the data pre-processing stage, (2) Invariance via Causal representation learning methods that are utilized during the representation learning stage, and (3) Invariance via Transferring Causal mechanisms methods that are applied during the classification stage of the pipeline. 

Furthermore, we present the different subcategories associated with each category (when applicable) and discuss the methods for each corresponding sub-category. A summary of the categorization can be seen in Fig.~\ref{categorization}.In the following sub-sections, we provide a comprehensive and detailed review of these methods corresponding to the above order and discuss their differences and theoretical connections.
\begin{figure}
    \centering
    \includegraphics[width=0.7\linewidth]{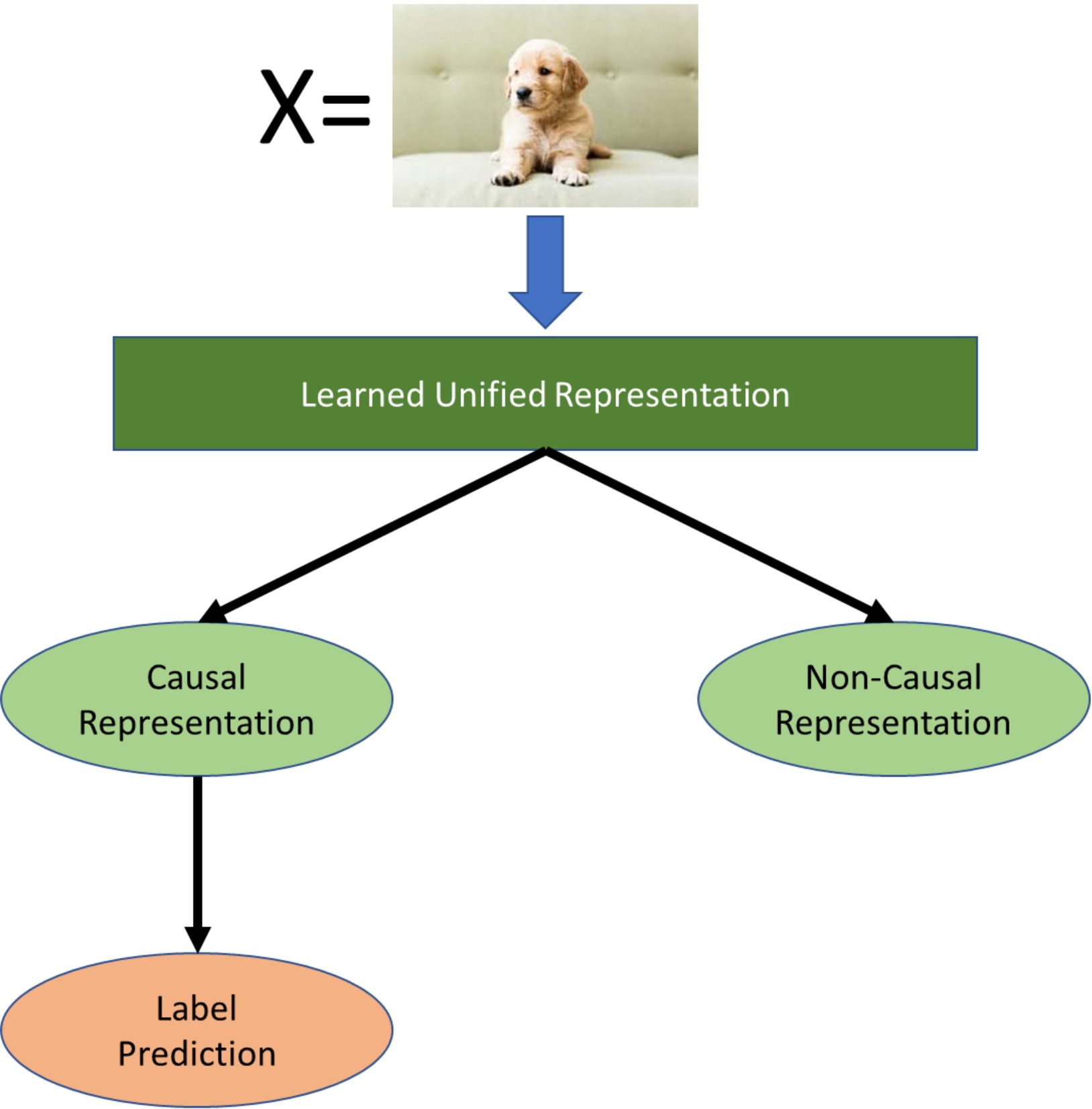}
    \caption{A simple causal graph that disentangles the input representations into causal and non-causal representations. Only the causal representations are used for prediction.}
    \label{simplecg}
\end{figure}

These frameworks learn generalizable feature representations across different domains. Often, these representations are interpreted as causal feature representations. One possible way to learn such causal representations is to disentangle the input representations into causal and non-causal feature representations. By doing so, the domain shift can be justified as interventions on the non-causal features. In such a condition, the primary goal is to minimize a loss that is robust under changes in the distribution of these style (non-causal) features. For example, in Fig.\ref{simplecg}, the learned input representations can be divided into causal and non-causal features. Utilizing only the causal features for the task can improve generalizability.

\subsection{Invariance via Causal Data Augmentation}
In this section, we present the frameworks that achieve invariance via causal data augmentation. Frameworks in this category utilize causal features and augment the data by considering all possible confounders/ spurious variables. Although these methods eventually aim to learn causal representations, the mechanisms they employ (such as identifying the features to augment on) follow a causal procedure thus promoting such methods to have their own category.

\subsubsection{Counterfactual Feature based Data Augmentation}
Zhang et al.~\cite{zhang2021learning1} propose a generalizable framework for the task of human pose estimation. The proposed framework aims to leverage generation of counterfactuals by intervening on the domain variable. In particular, these counterfactuals are generated by changing the domain variable. This framework enforces three main criteria to learn the causal representations: (1) Producing low-error prediction over the observed samples; (2) Producing low-error prediction over counterfactual samples; and (3) The counterfactual and observed feature representations are similar. The authors propose to implement the framework in two branches, namely observed branch and counterfactual branch. In the observed branch, images from source domain are fed to a feature extractor (encoder) to get the observed feature representation distribution. In the counterfactual branch, a GAN-based architecture is utilized to learn the distribution of counterfactuals from a ground-truth pose and random noise. Both observed and counterfactual representations are then fed to predictors to ensure they have high predictive power. To minimize the distance between observed and counterfactual representations a smooth-$l_1$ distance is utilized. The objective of this framework is given below:
\begin{equation}
\begin{aligned}
& \min _{\theta_f, \theta_h} \mathbb{E}_{(x,y,u)\sim(p(x),p(y),p(u)}\mathcal{L}_F(h(f(x)), y) \\
& \lambda_1\mathcal{L}_{CF}(h(g(u,y)), y) + \lambda_2\mathcal{L}_{dist}(f(x), g(u,y)),\\
\end{aligned}
\end{equation}
where $\mathcal{L}_F$ and $\mathcal{L}_{CF}$ denote the prediction loss over observed and counterfactual representations, $\lambda_1$ and $\lambda_2$ are hyperparameters.

Chen et al.~\cite{chen2021towards} propose the Interventional Emotion Recognition Network (IERN) to improve the visual emotion recognition framework's generalization via causal data augmentation. IERN first disentangles the input image into the context features (confounding features) and emotion features (causal features). Finally, the framework proposes to debias the classifier via backdoor criterion using the following objective:
\begin{equation}
\begin{aligned}
& \mathcal{L}_{cl} = \min_{f_c^\theta, g_e^\theta, f_b^\theta}(l_{CE}(\frac{1}{N_c}(\sum_{i=1}^{N_C} f_c(g_r(g_e(f_b(x)), C_i)), y_e))),
\end{aligned}
\end{equation}
where $f_c$ denotes the classifier, $g_e(f_b(x))$ represent the emotion features, $C_i$ are the confounders, ad $N_C$ is the number of confounders.
Ouyang et al.~\cite{ouyang2021causality}, propose a causal data augmentation approach to single-source domain generalization problem, where training data is only available from one source domain. The authors propose that the performance deterioration under domain shift may arise due to shifted domain-dependent features or shifted-correlation effect which is induced due to the presence of confounders. To deal with this problem, the authors propose to utilize causal intervention to augment the data and improve the robustness and generalization of the model. The proposed model consist of two parts: (1) global intensity non-linear augmentation (GIN) technique  that utilizes randomly-weighted shallow convolutional networks to  transforms images while keeping the shapes invariant; (2) interventional pseudo-correlation augmentation (IPA) technique that removes the confounder via re-sampling appearances of confounded objects independently. To augment the data, the framework first applies GIN on the input images to get new appearances. Then the two GIN-augmented of the same image are blended via IPA in a spatially-variable manner.
Mitrovic et al.~\cite{mitrovic2020representation} propose a novel self-supervised objective, Representation Learning via Invariant Causal Mechanisms (RELIC) based on a causal analysis of contranstive learning frameworks. To do so, the authors propose to model the data generation process similar to the causal graph presented in Fig.\ref{simplecg}.The graph indicates that the data is generated from content (causal) and style (non-causal) variables and that only the only content (causal) is informative of the downstream tasks (label prediction). Given the causal graph, the paper indicates that the optimal representation should be invariant predictor of proxy targets on correlated, not causally related features. Since none of the causal or correlational variables are known, the authors utilize data augmentations to simulate interventions on the styles (correlational) variables. Finally, the paper proposes a regularizer to enforce the invariance under data augmentations as follows:
\begin{equation}
\begin{aligned}
& \min \mathbb{E}_{X\sim p(X)}\mathbb{E}_{a_{lk}, a_{qt}\sim \mathcal{A}\times \mathcal{A}} \sum_{b \in\{a_{lk}, a_{qt}\}}\mathcal{L}_b(Y^R, f(X))\\
&\text{s.t.} KL(p^{do(a_{lk})}(Y^R\mid f(X)), p^{do(a_{qt})}(Y^R\mid f(X))) \leq \rho,
\end{aligned}
\end{equation}
where $\mathcal{A} = \{a_1, . . . , a_m\}$ is a set of data augmentations generated by intervening on the style variables, $\mathcal{L}$ is the proxy task loss and KL is the Kullback-Leibler (KL) divergence.
This KL-divergence based regularizer  enforces that the prediction of the proxy targets is invariant across data augmentations. 

\subsubsection{Gradient-based Data Augmentation}
Bai et al.~\cite{bai2021decaug} argue that many domain generalization methods work well for one dataset but perform poorly for others. They claim this happens because the domain generalization problems have multiple dimensions - correlation shift and diversity shift. Correlation shift is when the labels and the environments are correlated, and the relations change across different environments. Diversity shift means the data comes from different domains, thus having significantly different styles. For instance, the sketch of a horse, a cartoon horse, an image of a horse, and an art of a horse all represent horses in different styles. Furthermore, real-world data exists, which is a mixture of these shifts. To handle these dimensions simultaneously, they propose a novel decomposed feature representation and semantic augmentation approach. First, the proposed method decomposes the representations of the input image into context and category features. Then, they perform gradient-based semantic augmentation on context features, representing attributes, styles, and more; to disentangle the spurious correlation between features. The semantic data augmentation is performed by adversarially perturbing the feature space of the context related features of the original sample as follows:

\begin{equation}
\begin{aligned}
z_{i}^{c} = z_{i}^{c} + \alpha_i.\epsilon.\frac{\nabla_{z_{i}^{c}}(l(h_{\theta_{c}})(z_{i}^{c}, c_{i})))}{\|\nabla_{z_{i}^{c}}(l(h_{\theta_c})(z_{i}^{c}, c_{i}))) \|},
\end{aligned}
\end{equation}
where $z_i^c$ is the context feature representation, $h_{\theta_c}$ is the context feature discriminator, $\epsilon$ is a hyperparameter which controls the maximum
length of the augmentation vectors, and $\alpha_i$ is randomly sampled from $[0, 1]$. Unlike methods in the earlier category, this work does not aim to generate counterfactuals to improve generalization, rather they perform gradient based augmentation on disentangled context features
to eliminate distribution shifts for various generalization tasks.

\begin{figure}
    \centering
    \includegraphics[width=0.5\textwidth]{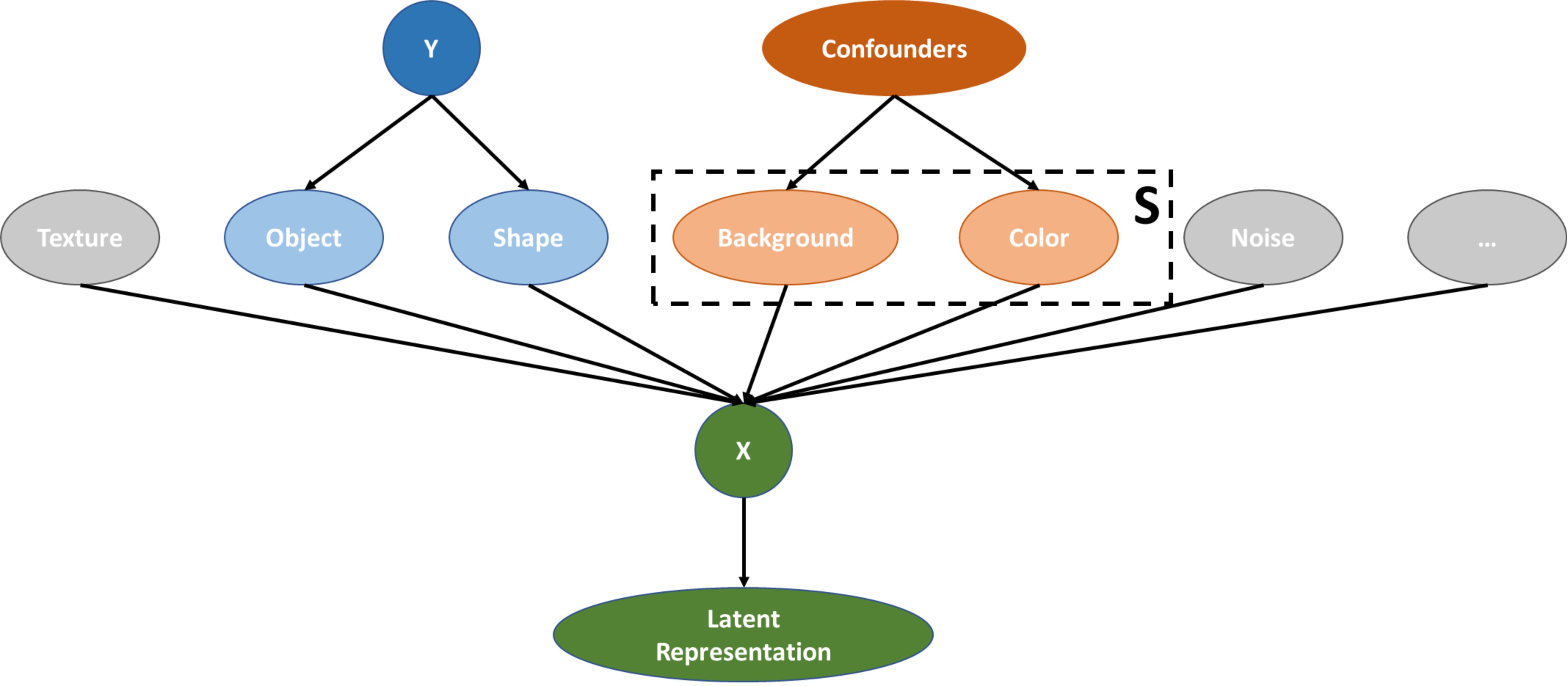}
    \caption{The data generating process as shown in ~\cite{chevalley2022invariant}. As per the authors, the input $X$ is generated by a collection of unobserved factors (such as texture, object, shape, and so on). Only $Y$, $X$, and possibly the confounders are observed during training time. $Y$ represents the variable that needs to be predicted. The generative factors are assumed to not have any causal relation among themselves. This work also assumes that the label and the confounders have an effect on the generative factors. $S$ represents the style variables.}
    \label{fig:invarchevally}
\end{figure}

\subsection{Invariance via Learning Causal Representations}
Causal DG methods aim to capture invariance with the aid of causal theories. Majority of these methods aim at learning a causal representation of the data that is highly predictive of the downstream task. For instance, consider the task of digit classification as discussed in the Introduction. Despite applying various transformations to the digits,(such as rotating the digits, coloring the digits, and so on) humans are still able to correctly associate digits with the labels. This phenomenon could be attributed to the fact that humans utilize the features that do not change during these transformations such as the shape of the digit. Thus, these features can be considered causal. However, learning these representations is not straightforward as these features are not directly observable making it challenging to guide the machine. One possible way to guide the model to rely on causal representations is with the aid of disentanglement. The range of works that consider disentanglement can be further divided into two parts namely, works that consider disentanglement with no causal interactions among the latent factors, and works that consider disentanglement with causal interactions among the latent factors. In this section we discuss these different categories.
\subsubsection{Disentanglement Assuming No Causal Interactions Among the Latent Factors}
This section discusses the works that assume the latent factors (i.e. the causal and non-causal factors) bare no causal interactions amongst each other. Majority of the papers utilize a simple causal graph as seen in Fig.~\ref{simplecg}. The input features are divided into causal and non-causal features.

\noindent\textbf{{Disentanglement when Auxiliary Variables are Available}}

A range of variables utilize auxiliary variables to aid the disentanglement process. For instance, in the task of image classification, auxiliary variables could be additional cues about the object, or they could be background variables indicating the image background. Thus, auxiliary variables can aid in distinguishing causal and non-causal features. The authors in~\cite{heinze2017conditional,trivedi2021contrastive,mahajan2021domain} aim to utilize auxiliary variables to separate causal from non-causal features, and learn the representations accordingly. For example, in grouped observations for certain datasets the same object (ID) is seen under multiple situations~\cite{heinze2017conditional}, which can guide the prediction to be based more on the latent core (causal) characteristics and less on the latent style (non-causal) features by penalizing between-object variance of the prediction less than variation for the same object. The authors of~\cite{heinze2017conditional} contend that direct interventions on style elements frequently result in the creation of a new domain. So, if $F_0$ represents the joint distribution of the $(ID, Y, X^{style})$ in the training distribution, then intervening on $X^{style}$ yields a new joint distribution of the $(\mathrm{ID}, Y, \tilde{X}^{style})$ indicated by $F$. As a result, we obtain the following class of distributions:
\begin{equation}
\mathcal{F}_{\xi}=\left\{F: D_{\text {style }}\left(F_{0}, F\right) \leq \xi\right\}
\end{equation}
where $D_{\text{style}}(F_{0}, F)$ is the distance between the two distributions. The primary goal is to optimize a worst-case loss over this distribution class. This loss can be formulated as, 
\begin{equation}
L_{\xi}(\theta)=\sup _{F \in \mathcal{F}_{\xi}} E_{F}\left[\ell\left(Y, f_{\theta}(X)\right)\right]
\label{cls_loss}
\end{equation}
For arbitrary strong interventions on the style features, the loss would be given by, 
\begin{equation}
L_{\infty}(\theta)=\lim _{\xi \rightarrow \infty} \sup _{F \in \mathcal{F}_{\xi}} E_{F}\left[\ell\left(Y, f_{\theta}(X)\right)\right]
\end{equation}
Minimizing this loss guarantees an accurate prediction that performs well even for significant shifts in the conditional distribution of style features. Rather than pooling over all examples, CoRe~\cite{heinze2017conditional} exploits the ID variable to penalize the loss function. The overall objective function is given by, 
\begin{equation}
\hat{\theta}^{\text {core }}(\lambda)=\operatorname{argmin}_{\theta} \hat{E}\left[\ell\left(Y, f_{\theta}(X)\right)\right]+\lambda \cdot \hat{C}_{\theta}
\end{equation}
where $\hat{C}_{\theta}$ is a conditional variance penalty of the form
\begin{equation}
\hat{C}_{f, \nu, \theta}:=\hat{E}[\widehat{\operatorname{Var}}(f_{\theta}(X) \mid Y, \text { ID })^{\nu}]
\end{equation}
for the conditional-variance-of-prediction, and 
\begin{equation}
\hat{C}_{\ell . \nu, \theta}:=\hat{E}\left[\widehat{\operatorname{Var}}\left(\ell\left(Y, f_{\theta}(X)\right) \mid Y, \mathrm{ID}\right)^{\nu}\right]
\end{equation}
for the conditional-variance-of-loss. $f_{\theta}(X)$ is the representation of the input $X$, $Y$ is the image label, $ID$ is the identifier label or the object label, and $\nu \in\{1 / 2,1\}$. Mahajan et al.~\cite{mahajan2021domain} argue that for representations, the class-conditional domain invariant objective is inadequate. They add to the CoRe framework by including an approach for when objects (ID variables) are not detected. When the stable feature distribution differs across domains, the class-conditional aim is insufficient to learn the stable features. To solve this issue, they use a causal graph to express within-class variance in stable features. The authors propose using the causal graph to learn the representation regardless of the classification loss.
Liu et al.~\cite{liu2021learning} claim that CoRe adds no new generative modeling efforts at the expense of limited capability for invariant causal processes They further contend that deep learning methods fail to generalize to unexplored domains because the representations they learnt blend semantic and stylistic information owing to false correlations. They propose using a causal generative model that follows a causal approach to describe the semantic and stylistic aspects independently to address these issues. Similar to CoRe, Makar et al.~\cite{makar2022causally} suggest that identifying invariant characteristics is a tough undertaking since it is impossible to distinguish the effect of non-causal factors without extra supervision. Instead of ID labels, the authors recommend employing auxiliary labels (such as picture background labels) to provide information about the irrelevant component that is available during training but not during testing. The authors offer a method for using these auxiliary labels to build a predictor whose risk is roughly invariant over a well-defined range of test distributions.

\begin{figure*}
    \centering
    \begin{subfigure}[t]{0.5\textwidth}
        \centering
        \includegraphics[width=0.6\linewidth]{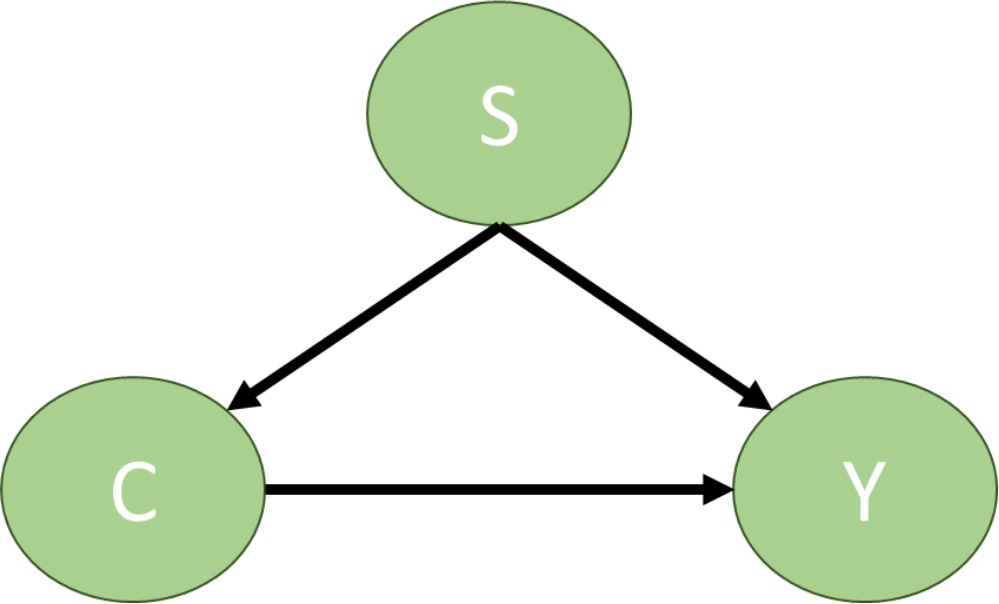}
        \caption{Causal graph for Backdoor adjustment.}
        \label{bda}
    \end{subfigure}%
    ~ 
    \begin{subfigure}[t]{0.5\textwidth}
        \centering
        \includegraphics[width=0.6\linewidth]{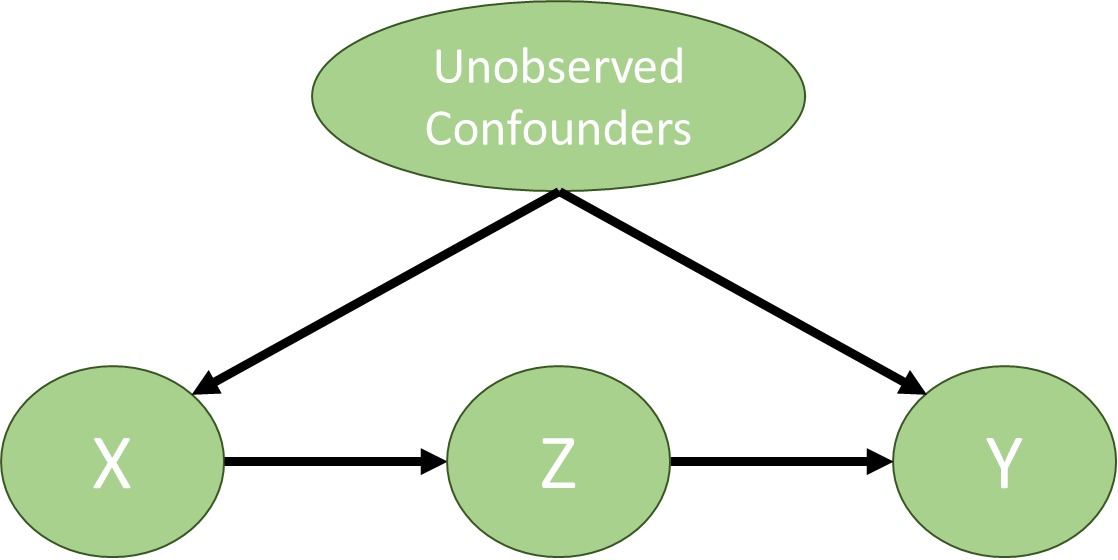}
        \caption{Causal graph for Front-door adjustment.}
        \label{fd}
    \end{subfigure}
\end{figure*}

Neet et al.~\cite{neet2022modeling} argue that causality-aware domain generalization frameworks impose regularization constraints to learn the invariance. However, when datasets with multi-attribute shifts are considered, deciding which regularization could lead to learning the invariance is difficult. To overcome this problem, the authors propose \textit{Causally Adaptive Constraint Minimization} (CACM). CACM, leverage the information provided by multiple independent shifts across attributes, assuming structural knowledge of the shifts. By identifying the correct constraints, CACM applies them as regularizers in the overall objective function. Moreover, the attributes correlate with the label that can change between train and test data. The model aims to learn a risk-invariant predictor that obtains minimum risk on all distributions. 
Mathematically,
\begin{equation}
\begin{aligned}
& g_{\text {rinv}} \in \arg \min _{g \in G_{rinv}} R_{P}(g) \forall P \in \mathcal{P} \\
& \text { where } G_{rinv}=\{g: R_{P}(g)=R_{P^{\prime}}(g) \forall P, P^{\prime} \in \mathcal{P}\},
\end{aligned}
\end{equation}
where $R$ is the risk of predictor $g$, and $\mathcal{P}$ represents all the distributions. 
The authors divide the auxiliary attributes $A$ into three types of attributes $\boldsymbol{A}_{\overline{ind}}$, which represents the attributes that are correlated with the label, $\boldsymbol{A}_{ind}$, the attributes that are independent of the label and $E$ denotes the domain. The shifts are based on the relationship between the auxiliary attributes $A$ and the label $Y$. Based on these distinctions of attributes, the authors define four kinds of shifts that are possible based on the causal graph. They are Independent, Causal, Confounded, and Selected. First, the authors identify the conditional independence constraints satisfied by the causal features and enforce that learned representation $\phi$ should follow the same constraint. The authors leverage the d-seperation~\cite{pearl2009causality} strategy and first, for every observed variable $V \in \mathcal{V}$ in the graph, check whether $\left(\boldsymbol{X}_{c}, V\right)$ are d-separated. If not, check whether $\left(\boldsymbol{X}_{c}, V\right)$ are d-separated conditioned on any subset of the remaining observed variables in $\mathcal{V} \backslash\{V\}$. Finally, the model applies those constraints as a regularizer to the standard ERM loss. Thus, the final objective function is formulated as,
\begin{equation}
g_{1}, \phi=\arg \min _{g_{1}, \phi} ; \quad \ell\left(g_{1}(\phi(\boldsymbol{x})), y\right)+\lambda^{*}(\text { RegPenalty }),
\end{equation}
where $\ell$ is the classification loss, $\lambda$ is a hyperparameter, and $RegPenalty$ is the regularization penalty. The penalty depends on the type of distribution shift for each attribute. 

\noindent\textbf{{Disentanglement when Auxiliary Variables are Unavailable}}

Although auxiliary variables can aid with causal disentanglement, these variables are not always easily available. A series of work aim to tackle the causal disentanglement problem in the absence of auxiliary variables~\cite{neet2022modeling,chen2021style,chevalley2022invariant,javed2020learning}. Chevalley et al.~\cite{chevalley2022invariant} argue that the invariant representations are reformulated as a feature of a causal process, and propose a regularizer that guarantees invariance via distribution matching. The proposed framework, in particular, describes the underlying generation process using the DAG shown in Fig.~\ref{fig:invarchevally}. Different domains are then represented in the provided DAG via soft-intervention on the domain variable. The invariant representations are therefore characterized as those that have no complete causal influence on the domain variable. The classification is formulated as:

\begin{equation}
\begin{split}
& \min_{Z=f(X)} \mathcal{L}(Y, c(Z))\\ 
&\text{s.t. } \mathbb{E}_{N_d, N'_d}[\operatorname{dist}(p^{do(d=N_d)(Z), p^{do(d=N'_d)(Z)}})],
\end{split}
\end{equation}
where $Z$ is the learned representation and dist denotes the distance between the distributions and $N_d$, $N'_d$ are the interventions on the domain variable. One way to disentangle causal and non-causal features in the presence of multiple domains, but the absence of auxiliary variables would be to utilize contrastive learning. The primary assumption under this setting is that the non-causal feature representations are similar for instances from the same domain. Thus, by guiding the machine learning model to learn non-causal representations, we can learn causal representations by learning orthogonal representations to the non-causal representations. In this setting, the objective function is usually represented as,
\begin{equation}
\mathcal{L}=\mathcal{L}^{cls}+\mathcal{L}^{con}
\end{equation}
where $\mathcal{L}^{cls}$ represent the classification loss, and $\mathcal{L}^{con}$ represents the contrastive loss. $\mathcal{L}^{cls}$ is formulated similar to Eq.~\ref{cls_loss} as it aims to predict the image label using the representation of the causal factors. $\mathcal{L}^{con}$ is formulated as, 
\begin{equation}
\mathcal{L}^{con}_{i,j} =-\log \frac{\exp (\operatorname{sim}(z_{i}, z_{j}) / \tau)}{\sum_{k=1}^{2 N} \exp (\operatorname{sim}(z_{i}, z_{k}) / \tau)}
\end{equation}
where $\tau$ is the temperature normalization factor, and $sim$ is the similarity function. The intuition here is that we want the similar representations $z_{i}$ and $z{j}$ to be close to each other, and the dissimilar representations $z_{i}$ and $z_{k}$ to be more distant from each other.
Recent works, including~\cite{chen2021style,trivedi2021contrastive} leverage this assumption to identify the causal features. Chen et al.~\cite{chen2021style} Concentrate on utilizing both semantic (causal) and stylistic (non-causal) characteristics. This paper's primary hypothesis is that instances from the same domain should exchange style information. This aids in the efficient disentanglement of style features, easing the search for actual semantic features with a low degree of freedom. Following the acquisition of style data, the network learns semantic information in orthogonal directions to the domain style. As a result, the network avoids overfitting style cues and instead concentrates on learning semantic elements. On the other hand, Trivedi et al.~\cite{trivedi2021contrastive} claim that standard domain generalization methods are inapplicable to computer games because games are more than just graphics, gaming images include functional qualities related with the game genre as well as aesthetic features unique to each game. As a result, representations derived from a pre-trained model perform well on the game on which it was trained, resulting in poor generalization. To overcome this issue, the authors recommend using contrastive learning. The authors use the causal graph to distinguish between content and stylistic aspects, stating that the game genre is defined by the content elements.

The ways mentioned earlier are helpful for the traditional machine learning setting. However, in an online learning scenario, where an agent continually learns as it interacts with the world, it is more challenging to identify the causal features. To tackle this problem, Javed et al.~\cite{javed2020learning} propose a framework for detecting and removing spurious features on a continuous basis. The framework's basic premise is that the association between a false characteristic and the goal is not continuous over time. The authors also contend that in order to infer a causal model from observable data, an agent must go through three phases. First, the agent must understand the basic data representations. Second, the agent must deduce the causal relationships between the variables. Finally, in order to generate correct predictions, the agent must grasp the interplay between these factors.

\subsubsection{Disentanglement Assuming Causal Interactions Among the Latent Factors}
This section discusses the works that assume the latent factors (i.e. the causal and non-causal factors) have a causal interaction amongst each other. Majority of the works assume the non-causal features to act as confounding factors and aim to leverage front-door or back-door criterion to mitigate confounding bias and improve generalization. There also exist works that aim to learn the nature of the relationships among the latent factors. For instance, Yang et al.~\cite{yang2021causalvae} suggest that when it comes to disentangling the generative elements of observational data, the bulk of frameworks presume that the latent factors are independent of one another. However, in practice, this assumption may not hold true. To solve this issue, they offer CausalVAE, a new framework. CausalVAE employs a Structural Causal Model layer that enables the model to recover latent components with semantics and structure using a Directed Acyclic Graph (DAG). First, the input $x$ is encoded to get the independent components $z$. The collected factors are subsequently processed by the Structural Causal Model (SCM) layer, which converts independent factors into causal endogenous ones. Finally, a masking method is employed to replicate the assignment operation of SCMs by propagating the influence of parental variables on their children.
Then the representation of the latent exogenous independent variables can be written as,
\begin{equation}
\epsilon=h(x, u)+\zeta
\end{equation}
where $\zeta$ is the independent noise term. After obtaining $\epsilon$, the causal representation is obtained as a linear SCM,
\begin{equation}
\mathbf{z}=\mathbf{A}^{T} \mathbf{z}+\epsilon=(I-\mathbf{A}^{T})^{-1} \epsilon
\end{equation}
where $\mathbf{A}$ represents the adjacency matrix of the causal graph, where $\mathbf{A}_{i} \in \mathbb{R}^{n}$ is the weight vector such that $\mathbf{A}_{ji}$ encodes the causal strength from $z_{j}$ and $z_{i}$, and $I$ is an identity matrix. Finally, the causal representations undergo a masking mechanism. This step resembles an SCM, which depicts how children are generated by their corresponding parental variables. The masking mechanism is formulated as,
\begin{equation}
z_{i}=g_{i}(\mathbf{A}_{i} \circ \mathbf{Z} ; \boldsymbol{\eta}_{i})+\epsilon_{i}
\end{equation}
where $\circ$ represents the element-wise multiplication, $\boldsymbol{\eta}_{i}$ is a parameter of the nonlinear function $g_{i}$. The final loss function consists of multiple terms, including the ELBO loss for VAE~\cite{kingma2013auto} and different constraints to ensure the latent variables $\mathbf{A}$ and $z$ are identifiable.

\noindent\textbf{{Mitigating Confounding Bias via Backdoor Adjustments}}

The backdoor adjustment aids in approximating the interventional distribution, which can guide in identifying the causal link between the causal features and the image label. Let $CS$ denote the representation of the causal features, $S$ denote the representation of the non-causal features, and $Y$ denote the image label. The causal-graph under this scenario can be visualized as shown in Fig.\ref{bda}. Similar to Eq.\ref{backdoor_real}, the backdoor adjustment can be formulated as,
\begin{equation}
P(Y \mid do(S=s_{k}))=\sum_{i=1}^{|CS|} P(Y \mid S=s_{k}, CS=c_{i}) P(CS=c_{i})
\label{backdoor}
\end{equation}

Zhang et al.~\cite{zhang2021learning} aim to utilize the causal graph to shed light on the poor generalization of classic person re-identification models when applied to unknown contexts. The primary premise of this research is that person pictures are influenced by two sets of latent random variables, namely identity-specific and domain-specific aspects. The authors propose Multi-Domain Disentangled Adversarial Neural Networks (MDANN), which learn two encoders from various datasets for embedding identity-specific (causal) and domain-specific (non-causal) components. To eliminate domain (identity) relevant information from embedded identity (domain) specific representations, the adversarial learning principle is used. Then, as shown in Eq.~\ref{backdoor}, a backdoor adjustment block (BA) is presented, which uses the identity-specific and domain-specific representations to achieve the approximation. The objective function is a combination of the backdoor adjustment and the classification loss.

Simiarly, Deng et al.~\cite{deng2021comprehensive} Attempt to use causality to enhance knowledge distillation among teacher-student models in order to improve generalizability. The authors create a causal graph to represent the causal linkages between the pre-trained instructor, the samples, and the prediction. To eliminate biased knowledge based on backdoor adjustment, the model employs the softened logits learnt by the teacher as the context information of an image. Overall, the proposed methodology captures full teacher representations while removing bias through causal intervention. Wang et al.~\cite{wang2021causal} present a causal attention model capable of distinguishing between causal and confounding picture aspects. The authors use backdoor adjustment to accomplish disentanglement of the causative and confounding aspects. To address the over-adjustment problem, they create data splits repeatedly and gradually self-annotate the confounders.

\noindent\textbf{{Mitigating Confounding Bias via Front-Door Adjustments}}

When the spurious features or confounders are unidentifiable or their distributions are hard to model, then one can use the front-door adjustment as it does not require explicitly modeling for the confounders. By introducing an intermediate variable $Z$ between the input $X$ and output $Y$, the front-door criterion transfers the requirement of modeling the intervening effects of confounders to modeling the intervening effects of the input. The motivation for this is clear if we consider a two state intervention. If we set the value of $X$, we can determine the corresponding value of $Z$, and we can then intervene again to fix that value of $Z$. By doing this for every value of $Z$ we are able to determine the effect of $X$ on $Y$. The causal-graph under this scenario can be visualized as shown in Fig.\ref{fd}. The front-door criterion can be formulated as shown in Eq.\ref{frontdoor}, where $x^{\prime}$ denotes the set of training data. By leveraging the front-door criterion, Li et al.~\cite{li2021confounder} present an approach for mitigating confounding bias in the absence of identifying the confounders The proposed technique simulates the interventions among various samples using the front-door criteria and then optimizes the global-scope intervening impact on the instance-level interventions. Unlike previous studies, this is the first effort to use the front-door criteria for learning causal visual cues by taking the intervention among samples into account.

\subsection{Invariance via Learning/Transfering Causal Mechanisms}
These papers train mechanisms (such as neural networks) invariant across different domains. The conditionals in these cases remain invariant. These frameworks impose that the invariance of conditionals is valid as long as the conditionals represent the causal mechanisms. Formally, given $\textbf{X}$ as the input, $Y$ as the label and $S^*$ as the subset of invariant predictors, the conditional distribution $P(Y^k \mid \mathbf{X}_{S^*}^k)$ is invariant across the $k$ domains.

\subsubsection{IRM and its Extensions}
When the training data originates from multiple domains, dividing the features into causal and non-causal features becomes challenging. One of the pioneering works - Invariant Risk Minimization (IRM)~\cite{arjovsky2019invariant}, was the first to address this problem. The authors propose to find a data representation such that a classifier trained on top of that representation matches for all domains. If the training data $D_{d}:=\{(x_{i}^{d}, y_{i}^{d})\}_{i=1}^{n_{d}}$ is collected under multiple domains $d \in \mathcal{E}_{\mathrm{tr}}$. Then, to achieve high generalizability, the model should minimize the following loss,
\begin{equation}
R^{\mathrm{OOD}}(f)=\max _{e \in \mathcal{E}_{\text {all }}} R^{e}(f)
\end{equation}
where $R^{e}(f):=\mathbb{E}_{X^{e}, Y^{e}}[\ell(f(X^{e}), Y^{e})]$  is the risk under domain $e$, $\mathcal{E}_{{all}}$ is a large set of unseen domains that are related to the source domains, $\ell(f(X^{e})$ is the representation function for the input $X$ from domain $e$, and $Y^{e}$ is the true image label. IRM is a learning paradigm to estimate data representations eliciting invariant predictors $w \circ \Phi$ across multiple domains. The constrained optimization problem that IRM aims to solve can be formulated as,
\begin{equation}
\begin{gathered}
\min _{\boldsymbol{w}, \Phi} \sum_{e \in \mathcal{E}_{{train}}} R^{e}(\boldsymbol{w} \circ \Phi) \\
\text { s.t. } \boldsymbol{w} \in \underset{\hat{\boldsymbol{w}}}{\operatorname{argmin}} R^{e}(\hat{\boldsymbol{w}} \circ \Phi),
\end{gathered}
\end{equation}
where $R^{e}$ is the cross-entropy loss for domain $e$, $\Phi$ is the feature extractor and $\boldsymbol{w}$ is a linear classifier. Since this is a bi-level optimization problem, it is difficult to optimize. By adopting the first-order approximation, the loss function is,
\begin{equation}
\min _{\Phi} \sum_{e \in \mathcal{E}_{train}} R^{e}(\Phi)+\lambda \cdot\|\nabla_{w \mid w=1.0} R^{e}(w \circ \Phi)\|,
\end{equation}
where $w \in \mathbb{R}$ is a dummy classifier. To find the invariant features across different domains, the authors assume that the data from all the environments share the same underlying Structural Equation Model (SEM).

Ahuja et al.~\cite{ahuja2021invariance} show that IRM fails in settings where invariant features capture all information about the label in the input. They further show that along with the information bottleneck constraints, the invariance principle works in both settings – when invariant features capture the label's information entirely and when they do not. Krueger et al.~\cite{krueger2021out} argue that the IRM principle is less effective when different domains are noisy. Furthermore, the authors also claim that IRM fails to achieve robustness w.r.t. covariate shifts. Also, the authors show that reducing differences in risk across training domains can reduce a model’s sensitivity to a wide range of extreme distributional shifts, including the challenging setting where the input contains causal and anti-causal elements. Since the proposed model seeks robustness to whichever forms of distributional shift, it is more focused on the domain generalization problem than IRM. Finally, the authors prove that equality of risks can be a sufficient criterion for discovering causal structure. Li et al.~\cite{li2022invariant} argue that the IRM principle fails for nonlinear classifiers and when pseudo-invariant features and geometric skews exist. To solve the problems of IRM, the authors in this work aim to utilize mutual information for causal prediction. Furthermore, they aim to adopt the variational formulation of the mutual information for nonlinear classifiers to develop a tractable loss function. The proposed model seeks to minimize invariant risks while at the same time mitigating the impact of pseudo-invariant features and geometric skews. Guo et al.~\cite{guo2021out} argue that the IRM principle guarantees the existence of an invariant optimal classifier for a set of overlapping feature representations across domains. However, as DNNs tend to learn shortcuts, they can circumvent IRM by learning non-overlapping representations for different domains. Thus, IRM fails when the spurious correlations are stronger than the invariant relations. The authors propose to utilize conditional distribution matching to overcome this problem. 

Huh et al.~\cite{huh2022missing} propose that by conserving the class-conditioned feature expectation $\mathbb{E}_{e}[f(x) \mid y]$ across the different domains, one could overcome the flaws in IRM. Bellot et al.~\cite{bellot2020accounting} argue that earlier works that optimize for worst-case error under interventions on observed variables are typically not optimal under moderate interventions, especially in the presence of unobserved confounders. They also argue that minimum average error solutions can optimize for any dependency in the data, and their performance deteriorates when the data is subjected to interventions on observed variables; but better adjusts to interventions on the confounders. Thus, causal and minimum average error solutions can be interpreted as two extremes of a distributionally robust optimization problem with a range of intermediate solutions that may have a more desirable performance. To this end, the authors propose Derivative Invariant Risk Minimization (DIRM) that interpolates between the causal solution and the minimum average error solution.

The IRM algorithm has also been effectively utilized in various applications. For instance, Francis et al.~\cite{francis2021towards} highlights the application of generalization in federated learning models. The authors argue that the current federated learning models have poor generalizability as most of the data for the federated server are not i.i.d making it difficult for these models to generalize better. To this end, they propose to leverage causal learning to improve the generalizability in a federated learning scenario. The proposed model consists of client and global server layers. The client layer extracts the features from their respective input data. The global layer aids the participating clients in exchanging intermediate training components and trains the federated model in collaboration by minimizing the empirical average loss. The authors leverage IRM to learn invariant predictors in a federated learning setup that can attain an optimal empirical risk on all the participating client domains.

\subsubsection{Utilizing Auxiliary Functions to Model Conditional Distributions}
Generally, the domain generalization problem can be solved by posing it from a causal discovery point of view. Muller et al.~\cite{muller2021learning} argue that when posed from a causal discovery viewpoint, a set of features exists for which the relation between this set and the label is invariant across all domains. They claim that this happens because of the Independent Causal Mechanisms (ICM) principle, which states that every mechanism acts independently of the others. Given a joint distribution of the input $\mathbf{X}$, the chain rule can decompose this distribution into a product of conditionals. This can be formulated as,
\begin{equation}
p_{\mathbf{X}} (x_{1}, \ldots, x_{D})=\prod_{i=1}^{D} p_{i} (x_{i} \mid \mathbf{x}_{pa(i)}),
\end{equation}
where $\mathbf{x}_{pa(i)}$ refer to the causal parents of $x_{i}$, and the  conditionals $p_{i}$ of this causal factorization are called \textit{causal mechanisms}. The normalizing flows model complex distributions using invertible functions $T$, which map the densities of interest to latent normal distributions. Thus, the authors represent the conditional distribution $P(Y \mid h(\mathbf{X}))$ using a conditional normalizing flow. They seek to learn a mapping $R=T(Y ; h(\mathbf{X}))$ such that, $R \sim \mathcal{N}(0,1) \perp h(\mathbf{X})$ when $Y \sim P(Y \mid h(\mathbf{X})$. $T$ can be learned by minimizing the negative log-likelihood as,
\begin{equation}
\begin{aligned}
\mathcal{L}_{\mathrm{NLL}}(T, h):=& \mathbb{E}_{h(\mathbf{X}), Y}[\| T(Y ; h(\mathbf{X}) \|^{2} / 2\\
&.-\log |\operatorname{det} \nabla_{y} T(Y ; h(\mathbf{X}))|]+C,
\end{aligned}
\end{equation}
where $C$ is a constant, $\operatorname{det} \nabla_{y}$ is the Jacobian  determinant.
Finally, the authors propose a differentiable two-part objective function formulated by,
\begin{equation}
\arg \min _{\theta, \phi} (\max _{e \in \mathcal{E}}\{\mathcal{L}_{\mathrm{NLL}}(T_{\theta}, h_{\phi})\}+\lambda_{I} \mathcal{L}_{I}(P_{R}, P_{h_{\phi}(\mathbf{X}), E})),
\end{equation}
where $\phi$ and $\theta$ denote model parameters, $h_{\phi}$ denotes the feature extractor, $E$ denotes the domain, and $\mathcal{L}_{I}$ denotes the Hilbert Schmidt Independence Criterion (HSIC)~\cite{gretton2005measuring}, a kernel-based independence measure that penalizes dependence between the distributions of $R$ and $(h_{\phi}(\mathbf{X}), E)$. The distribution loss stems from the theorem that if $R \perp(h(\mathbf{X}), E)$. Then, it holds that $Y \perp E \mid h(\mathbf{X})$.

\subsubsection{Graphical Criterion based Methods}
Zheng et al.~\cite{zheng2021learning} leverage the causal factorization from the ICM principle to identify the autonomous generating factors (i.e., distribution of each variable given its cause), as changing one will not affect others. Due to this autonomy, it is necessary to have varied factors to account for the distributional shifts. The authors classify this set of variables as the mutable set, which is not assumed to be known. The authors further argue that obtaining a set of invariant predictors is possible, conditioned on the intervened mutable variables and any stable subset. However, it becomes a challenging task when the degeneration condition does not hold. The degeneration condition implies that the predictor with the whole stable set is min-max optimal if the intervened distribution can degenerate to the conditional one. When this condition is determined not to hold, the authors transform the worst-case quadratic loss into an optimization problem over all subsets of the stable set, the minimizer of which is sufficient and necessary for the predictor to be min-max optimal.

Lv et al.~\cite{lv2022causality} propose a novel method, Causality Inspired Representation Learning (CIRL), which aims to learn causal representations. They argue that the intrinsic causal mechanisms (formalized as conditional distributions) can be feasible to construct if the causal factors are given. However, it is hard to recover the causal factors since they are unobservable. To alleviate this problem, they propose to learn representations based on three general properties of the causal factors. They are (1) The causal factors are separated from the non-causal factors; (2) The factorization of the causal factors should be jointly independent; and (3) the factors should be causally sufficient, i.e., they should contain the necessary causal information needed for the classification task. These properties are based on Common Cause Principle and Independent Causal Mechanism principle. First, CIRL performs interventions using Fourier transforms to differentiate between causal and non-causal features. Next, representations of the augmented and the original images are learned as $r=\hat{g}(x)$, where $g(.)$ is the representation function. To simulate the causal factors that remain invariant to the intervention, the model enforces $\hat{g}$ to keep
unchanged dimension-wisely, as,
\begin{equation}
\max _{\hat{g}} \frac{1}{N} \sum_{i=1}^{N} COR(\tilde{\boldsymbol{r}}_{i}^{o}, \tilde{\boldsymbol{r}}_{i}^{a}),
\end{equation}
where $\tilde{\boldsymbol{r}}_{i}^{o}$ and $\tilde{\boldsymbol{r}}_{i}^{a}$ are the normalized Z-score of the $i^{th}$ column of $\mathbf{R}^{o}=[(\boldsymbol{r}_{1}^{o})^{T}, \ldots,(\boldsymbol{r}_{B}^{o})^{T}]^{T} \in \mathbb{R}^{B \times N}$ and $\mathbf{R}^{a}=$ $[(\boldsymbol{r}_{1}^{a})^{T}, \ldots,(\boldsymbol{r}_{B}^{a})^{T}]^{T}$, respectively, $COR$ is a correlation function, and $N$ is the number of dimensions. The joint independence among the causal factors is imposed by ensuring that any two dimensions of the representations are independent. Thus, the same dimension of $\mathbf{R}^{o}$ and $\mathbf{R}^{a}$ are considered positive pairs. In contrast, the different dimensions are considered negative pairs. Positive pairs should hold a stronger correlation when compared to negative pairs. This can be formulated with a factorization loss $\mathcal{L}_{Fac}$ which can be formulated as,
\begin{equation}
\mathcal{L}_{Fac}=\frac{1}{2}\|\boldsymbol{C}-\boldsymbol{I}\|_{F}^{2},
\end{equation}
where $C$ is the correlation matrix computed as,

$\boldsymbol{C}_{i j}=\frac{\langle\tilde{\boldsymbol{r}}_{i}^{o}, \tilde{\boldsymbol{r}}_{j}^{a}\rangle}{\|\tilde{\boldsymbol{r}}_{i}^{o}\|\|\tilde{\boldsymbol{r}}_{j}^{a}\|}, i, j \in 1,2, \ldots, N$, and $I$ is the identity matrix. Classification loss with the causal representations can be computed to ensure the representations are causally sufficient. However, inferior dimensions may carry relatively less causal information and make a small contribution to classification. Thus, the authors design an adversarial mask module to detect the inferior dimensions. The module learns the importance of the different dimensions. The dimensions corresponding to the largest $\kappa \in(0,1)$ ratio are regarded as superior. In contrast, the rest are regarded as inferior ones. The authors use the GumbelSoftmax trick [16] to sample a mask with $\kappa$N values approaching 1. Thus, the classification loss is given by, 
\begin{equation}
\begin{aligned}
&\mathcal{L}_{cls}^{sup}=\ell(\hat{h}_{1}(r^{o} \odot m^{o}), y)+\ell(\hat{h}_{1}(r^{a} \odot m^{a}), y) \\
&\mathcal{L}_{cls}^{inf}=\ell(\hat{h}_{2}(r^{o} \odot(1-m^{o})), y)+\ell(\hat{h}_{2}(r^{a} \odot(1-m^{a})), y).
\end{aligned}
\end{equation}
The final objective function is represented as, 
\begin{equation}
\min _{\hat{g}, \hat{h}_{1}, \hat{h}_{2}} \mathcal{L}_{cls}^{sup}+\mathcal{L}_{cls}^{inf}+\tau \mathcal{L}_{Fac}, \quad \min _{\hat{w}} \mathcal{L}_{cls}^{sup}-\mathcal{L}_{cls}^{inf},
\end{equation}
where $\tau$ is the trade-off parameter. Similarly, Wang et al.~\cite{wang2021contrastive} argue that rather than the distribution of the features, it is the causal mechanism that remains invariant across domains. By utilizing feature-level invariance with the aid of regularizers, the model may learn spurious features that are only correlated to the label. Thus, to this end, the authors propose to treat the machine learning models as SCMs. In other words, the authors propose a novel constraint that measures the Average Causal Effect (ACE) between the causal attributions in the networks. They leverage contrastive learning and introduce ACE contrastive loss that pairs samples from the same class and different domains as positive pairs and those of different classes as negative pairs. The ACE contrastive loss regularizes the learning procedure and encourages domain-independent attributions of extracted features. The causal attribution of the neuron $z^j$, corresponding to the $j^{th}$ feature, on the output $y$ is calculated as the subtraction between the interventional expectation of $y$ when $z^j=\alpha$ and a baseline of $z^j$, given by,
\begin{equation}
c_{do(z^{j}=\alpha)}^{y}=\mathbb{E}[y \mid do(z^{j}=\alpha)]-\mathbb{E}_{z^{j}}[\mathbb{E}[y \mid do(z^{j}=\alpha)]].
\end{equation}
The ACE vector for the $i^th$ samples is given by,
\begin{equation}
\boldsymbol{c}_{i}=[c_{i}^{1}, c_{i}^{2}, \ldots c_{i}^{n}].
\end{equation}
The final objective function is then formulated as,
\begin{equation}
\mathcal{L}_{\mathcal{A}}(\mathbf{D} ; \theta, \phi)=\sum_{i=1}^{m} \ell(g_{\phi}(f_{\theta}(x_{i})), y_{i})+\rho \sum_{i=1}^{m} \mathcal{A}_{\theta, \phi}(x_{i}),
\end{equation}
where the first term denotes the classification loss, $g_{phi}$ is the classifier and $f_{\theta}$ are the features and $\mathcal{A}_{\theta, \phi}(x_{i})$ represents the Contrastive ACE loss which is given by,
\begin{equation}
\max \{\operatorname{dist}(\mathbf{c}_{i}, \mathbf{c}_{q(\mathcal{P}_{i})})-\operatorname{dist}(\mathbf{c}_{i}, \mathbf{c}_{q(\mathcal{N}_{i})})+\delta, 0\},
\end{equation} 
where $q(\mathcal{P}_{i})$ is an index sampled uniformly at random from
the positive pairs' set,  $q(\mathcal{N}_{i})$ is from the negative pair, $\delta$ is a margin variable, $dist$ is a distance metric. Sun et al.~\cite{sun2020latent} argue that when dealing with sensory-level data such as modeling pixels, it is beneficial to model the problem similar to human perception; i.e., the causal factors of the label $Y$ is related to unobserved abstractions $S$ via a mechanism $f_{y}$ such that $Y \leftarrow f_{y}(S, \varepsilon_{y})$, where $\varepsilon$ is a noise term. At the same time, there exist latent variables $Z$ that, along with variables $S$ generate the input image $X$ via mechanism $f_{x}$ such that $X \leftarrow f_{x}(S, Z, \varepsilon_{x})$. Under this situation, domain shifts occur when variables $Z$ are allowed to correlate to the variables $S$ spuriously. For instance, when dealing with the image classification problem, the background features can be classified as $Z$, and the object-related abstractions such as shape can be classified as $S$. The authors encapsulate this information in a set of causal models. They argue that the generating mechanisms $f_{x}$ and $f_{y}$ are invariant across domains. At the same time, the spurious relation between $Z$ and $S$ is allowed to vary. Mathematically, Causal Invariance refers to the condition when $P(Y \mid do(s))$ and $P(X \mid do(s), do(z))$ are stable to the shift across domains. The authors finally reformulate the Variational Bayesian method to estimate the Causal Invariance during training and optimize it during testing.

\subsubsection{Kernel-based Optimization Methods}
Muandet et al.~\cite{muandet2013domain} was one of the earliest works aiming at enhancing the generalizability of the machine learning models from a causal perspective. The authors argue that the conditional distribution of the label $Y$, given an input $X$ is stable. However, the marginal distribution i.e., $P(X)$ may fluctuate smoothly. Due to this fluctuation machine learning models may suffer from model misspecification, i.e. the model fails to account for everything that it should. To alleviate this problem, the authors propose Domain Invariant Component Analysis (DICA). DICA aims to find data transformations that minimize the difference between the marginal distribution of different domains while preserving the stable conditional $P(Y \mid X)$. They introduce a domain generalization approach that learns an invariant transformation across domains between inputs and outputs by minimizing the dissimilarities between different domains. Basically, the objective of this work is to find a transformation that satisfies the following two properties: (1) Minimizing the distance between the distribution of the samples transformed via this transformation; and (2) The learned transformation between input and output remains invariant across different domains.
To do so, a kernel-based optimization objective is defined as:
\begin{equation}
\begin{aligned}
\label{eq:finalinver}
&\max_{B\in \mathbb{R}^N\times M} \frac{\frac{1}{n}\operatorname{Tr}(B^TL(L+n\epsilon I_n)^{-1}K^2B}{\operatorname{Tr}(B^TKQKB+BKB)}, \\
\end{aligned}
\end{equation}
where K and Q are the block kernel and coefficient matrices, respectively and B is the estimator which satisfies the two desired properties.

\section{Domain Generalization in Text and Graphs}
\label{cat_tg}
So far we have surveyed existing works on the causal domain generalization for the image data. In the following we discuss the frameworks designed for text and graph data.
\subsection{Invariance Learning for Text}
With the advances of large pre-trained models, Natural Language Processing models have gained widespread success over multiple applications in the real world. However, these models are brittle to out-of-domain samples~\cite{hendrycks2020pretrained}.  A series of works, showcase how the language models rely on spurious correlations for classification. For instance, Wang et al.~\cite{wang2020identifying} show that words such as \textit{Spielberg} is correlated to positive movie reviews. Wulczyn et al.~\cite{wulczyn2017ex} show that a toxicity classifier learns that “gay” is correlated with toxic comments. To address the reliance on spurious correlations recent works~\cite{choi2022c2l,wang2021robustness,veitch2021counterfactual,peyrard2021invariant,cheng2022bias} aim at leveraging different causal techniques to improve the generalizability of the NLP models. Interested readers can refer to~\cite{feder2021causal} which focuses on how to infer causal relations in natural language processing models. 

Causality-aware domain generalization works in the natural language domain are also aligned with our defined categories. For instance, recent works in NLP~\cite{srivastava2020robustness,kaushik2019learning} leveraged human in-the-loop systems to make the models robust to spurious correlations by leveraging human common sense of causality. They augment training data with crowd-sourced annotations about reasoning of possible shifts in unmeasured variables and finally conduct robust optimization to control worst-case loss. While these works highlight that human annotations improve the robustness, collecting such annotations can be costly. To overcome this problem Wang et al.~\cite{wang2021robustness} propose to train a robust classifier with automatically generated counterfactual samples which is aligned with causal data augmentation approaches as listed earlier in the categorization section. The authors aim at utilizing the closest opposite matching approach to identify the likely causal features, and then generate counterfactual training samples by substituting causal features with their antonyms and assigning opposite labels to the newly generated samples. 

Mathematically, for a given sample $(d, y)$, where $d$ represents document and $y$ represents the label, the corresponding counterfactual $(d', y')$ is generated by (i) substituting causal terms in $d$ with their antonyms to get $d'$ , and (ii) assigning an opposite label $y'$ to $d'$. The authors propose a two-stage process  as follows:
\begin{itemize}
\item Using models such as logistic regression, the authors identify the strongly correlated terms $\left\langle t_{1} \ldots t_{k}\right\rangle$ as  candidate causal features.
\item For each top term $t$ and a set of documents containing $t: D_{t}=\langle d_{1} \ldots d_{n}\rangle$, the authors utilize context similarity to identify documents with \textit{opposite} labels.
\item Then, the authors select the highest score match for each term and identify likely causal features by picking those whose closest opposite matches have scores greater than a threshold (0.95 is used below). 
\item Then for each sample, the authors substitute the causal terms with antonyms and generate counterfactual samples. Finally, they train the robust classifier using the original and counterfactual data.
\end{itemize}

Similarly, Tan et al.~\cite{tan2021causal} propose to construct meaningful counterfactuals that would reflect the model’s decision boundaries. The authors leverage rule-based schemes that negate causal relations or strengthen conditionally causal sentences.

Choi et al.~\cite{choi2022c2l} aim to utilize contrastive learning to enhance the representations of the causal features, this is aligned with capturing invariance via learning causal representations. The porposed model $C^{2}L$, first aims to identify the causal tokens based on attribution scores. Formally, to identify the important tokens, the authors leverage attribution scores as follows:
\begin{equation}
g_{i}=\|\nabla_{\mathbf{w}_{i}^{p}} \mathcal{L}_{\text {task }}(x, y ; \phi)\|^{2},
\end{equation}
where $x$ denotes the input, $y$ denotes the label, $g_{i}$ denotes the gradient magnitude computed from the classifier $f_{\phi}$, and $\mathcal{L}_{task}$ denotes the cross-entropy loss. The gradient-based score of token $w$ is aggregated over all the training texts having the token $w$ by:
\begin{equation}
s^{grad}(w)=\sum_{(x, y) \in \mathcal{D}} \frac{1}{n_{w, x}} \sum_{i \in\{1, \ldots, T\}} \mathbb{I}(w_{i}=w) \cdot g_{i}
\end{equation}
where $\mathbb{I}$ is an indicator function, and $n_{w, x}$ is the number of word $w$ in the input $x$. After obtaining the scores for each tokens, the authors employ a causal validation technique to identify the causal tokens. As per this step, the authors compute the Individual Treatment Effect (ITE) of each token on the label. The main intuition behind this step is that, if the masked text can be reconstructed into multiple examples with different classes, we can decide the masked term has a causal effect. To this end, the authors use BERT with dropout mechanism to identify the top-$k$ substitutions for the token $w$. The $k$ candidates are then passed through the classifier to obtained the predicted labels $\hat{y}$. By testing whether the $k$ labels are evenly distributed into the classes, we can decide the high-attributed token $w$ as causal to its task label $y$. 

Finally, the authors leverage contrastive learning to better learn the causal structure of the classification task. After obtaining the causal features, the authors generate causal triplets of the form $\left(x, x^{+}, x^{-}\right)$. $x^{-}$ denotes the counterfactual pair that is generated by masking out causal words. In contrast, $x^{+}$ denotes the factual pair generated by masking one of the non-causal words that is still recognized as the original label $y$, which helps to learn a model invariant to these features. The contrastive objective aims at mapping the representation of $x$ closer to $x^{+}$ and further from $x^{-}$. The objective can be formulated as,
\begin{equation}
\begin{aligned}
&\mathcal{L}_{c}(x ; \theta)=\max (0, \\
& \Delta_{m}+\frac{1}{J} \sum_{j=1}^{J} s_{\theta}(x, x_{j}^{+})-\frac{1}{J} \sum_{j=1}^{J} s_{\theta}(x, x_{j}^{-}))
\end{aligned}
\end{equation}
where $J$ is the number of positive/negative pairs, $\Delta_{m}$ is a margin value and $s_{\theta}(\cdot, \cdot)$ is distance between the representations. The final objective function is given by,
\begin{equation}
\mathcal{L}=\mathcal{L}_{\text {task }}+\lambda \mathcal{L}_{c}
\end{equation}
where $\lambda$ is a balancing coefficient for the contrastive objective.

Inspired by IRM~\cite{arjovsky2019invariant}, Peyrard et al.~\cite{peyrard2021invariant} propose invariant Language Modeling (iLM), a framework that generalizes better across multiple environments for language models in NLP. iLM is aligned with learning invariance by transferring causal mechanisms. The ILM aims at utilizing a game-theoretic implementation of IRM for language models. In this case, the invariance is achieved by a specific training schedule in which each environment competes with the others to optimize their environment-specific loss by updating subsets of the Language Model (LM) in a round-robin fashion.

The IRM-games~\cite{ahuja2020invariant} method aims to change the training procedure of IRM by using a game-theoretic perspective in which each environment $e$ is tied to its own classifier $w^{e}$, and the feature representation $\phi$ is shared. The global classifier $w$ is defined as an ensemble formulated as, 
\begin{equation}
 w=\frac{1}{|\mathscr{E}|} \sum_{e \in \mathscr{E}} w^{e}
\end{equation}
where $\mathscr{E}$ represents the set of training environments. 
Each environment takes turns to minimize their own empirical risk $R^{e}(w \circ \phi)$ w.r.t their own classifier $w^{e}$, while the shared $\phi$ is updated periodically. The authors adopt this IRM-games setup for language models. First, the shared representation can be represented by the main body of the encoder of an LM, and $w^{e}$ is the language modeling head that outputs the logits after the last layer. The multiple environments for this task can be the different sources from which text data emerges, for instance, encyclopedic texts, Twitter, news articles, and so on. Suppose for each environment the data can be represented as $\{(X^{e}, Y^{e})\}_{e=1 \ldots n}$. A forward pass on a batch $(x_{i}, y_{i})$ sampled according to $P(X^{i}, Y^{i})$ from environment $i$ involves $n$ language modeling heads $\{w_{e}\}_{e=1 \ldots n}$:
\begin{equation}
\hat{y}=\operatorname{softmax}(\sum_{e=1}^{n} w_{e} \circ \phi(x_{i}))
\end{equation}
Based on the task, a modeling loss $\mathcal{L}$ can be applied to the output $\hat{y}$. Since the underlying causal model is not known for language models, utilizing the data stemming from different environments can facilitate in learning the invariant relationships. However, the choice of environments is an important step as the environments define which relations are spurious in nature.
\subsection{Invariance Learning for Graphs}
Unlike vision and natural language data, graph data is heterogeneous. Graph Neural Networks (GNNs) fuse heterogeneous information from node features and graph structures to learn effective node embeddings. However, the complex and unobserved non-linear dependencies among representations are much more difficult to be measured and eliminated than the linear cases for decorrelation of non-graph data. In out-of-distribution scenarios, when complex heterogeneous distribution shifts exist, the performance of current GNN models can degrade substantially, mainly induced by spurious correlations. The spurious correlations intrinsically come from the subtle correlations between irrelevant and relevant representations. To address this problem, a variety of methods have been proposed. 

Liu et al.~\cite{liu2022graph} leverage graph data augmentations to identify graph rationales. The authors aim to augment the rationale subgraph by removing its environment subgraph and combining it with different environment subgraphs. Furthermore, they propose a framework, namely, GREA, that leverages masking to separate the rationales from the environment. After learning the node representations with the aid of a GNN, a Multi-Layer Perceptron (MLP) is used to map the node representations to a mask vector $\mathbf{m} \in(0,1)^{N}$ on the nodes in the set $\mathcal{V}$. $m_{v}=\operatorname{Pr}\left(v \in \mathcal{V}^{(r)}\right)$ is the node-level mask that indicates the probability of node $v \in \mathcal{V}$ being classified into the rationale subgraph. It is formulated as,
\begin{equation}
m=\sigma\left(\mathrm{MLP}_{1}\left(\mathrm{GNN}_{1}(g)\right)\right)
\end{equation} 
GREA uses another GNN encoder to generate contextualized node representations $\mathbf{H}: \mathbf{H}=\mathrm{GNN}_{2}(g)$. With $\mathbf{m}$ and $\mathbf{H}$, the rationale subgraph and environment subgraph can be easily separated in the latent space. The rationale and environment subgraph are generated as,
$$
\mathbf{h}^{(r)}=\mathbf{1}_{N}^{\top} \cdot(\mathrm{m} \times \mathbf{H}), \quad \mathbf{h}^{(e)}=1_{N}^{\top} \cdot\left(\left(1_{N}-\mathbf{m}\right) \times \mathbf{H}\right),
$$
where $\mathbf{1}_{N}$ denotes the $N$-size column vector with all entries as 1, and $\mathbf{h}^{(r)}, \mathbf{h}^{(e)} \in \mathbb{R}^{d}$ are the representation vectors of rationale graph $g^{(r)}$ and environment graph $g^{(e)}$, respectively. After the rationale and environment separation, the model leverages two augmentation strategies to make predictions. First, it combines each rationale subgraph with multiple environment subgraphs to generate augmented samples which can improve the model's robustness and generalization. The prediction is made as,
\begin{equation}
\hat{y}_{(i, j)}=\operatorname{MLP}_{2}\left(\mathbf{h}_{(i, j)}\right)
\end{equation}
where $\hat{y}_{i,j}$ is computed as,
\begin{equation}
\mathbf{h}_{(i, j)}=\operatorname{AGG}\left(\mathbf{h}_{i}^{(r)}, \mathbf{h}_{j}^{(e)}\right)=\mathbf{h}_{i}^{(r)}+\mathbf{h}_{j}^{(e)}
\end{equation}
Second, it removes the environment subgraph and uses only the rationale subgraphs to make predictions as follows, 
\begin{equation}
\hat{y}_{i}^{(r)}=\operatorname{MLP}_{2}\left(\mathbf{h}_{i}^{(r)}\right)
\end{equation}

For instance, Fan et al.~\cite{fan2021generalizing} propose to leverage causality to overcome the subgraph-level spurious correlations to improve the GNN generalizability. They analyzed the degeneration of GNNs from a causal view and propose a novel causal variable distinguishing regularizer to decorrelate each high-level variable pair by learning a set of sample weights. The sample reweighting method aids in eliminating the dependence between high-level variables, where non-linear dependence is measured by weighted Hilbert-Schmidt Independence Criterion (HSIC)~\cite{gretton2005measuring}. HSIC is formulated as,
\begin{equation}
\operatorname{HSIC}_{0}^{k, l}(U, V, \mathbf{w})=(m-1)^{-2} \operatorname{tr}(\hat{\mathbf{K}} \mathbf{P} \hat{\mathbf{L}} \mathbf{P})
\end{equation}
where $\hat{\mathbf{K}}, \hat{\mathbf{L}} \in \mathbb{R}^{m \times m}$ are weighted RBF kernel matrices containing entries $\hat{\mathbf{K}}_{i j}=k\left(\hat{U}_{i}, \hat{U}_{j}\right)$ and $\hat{\mathbf{L}}_{i j}=l\left(\hat{V}_{i}, \hat{V}_{j}\right)$, $\hat{U}=\left(\mathbf{w} \cdot \mathbf{1}^{\mathrm{T}}\right) \odot U$, and $\hat{V}=\left(\mathbf{w} \cdot \mathbf{1}^{\mathrm{T}}\right) \odot V$. Finally, the weights are optimized as,
\begin{equation}
\mathbf{w}^{*}=\underset{\mathbf{w} \in \Delta_{m}}{\arg \min } \sum_{1<p<n_{L}} \mathrm{HSIC}_{0}^{k, l}\left(\mathbf{H}_{, 0: d}, \mathbf{H}_{,(p-1) d: p d}, \mathbf{w}\right),
\end{equation}
where $\Delta_{m}=\left\{\mathbf{w} \in \mathbb{R}_{+}^{n} \mid \sum_{i=1}^{m} \mathbf{w}_{i}=m\right\}$,  and we utilize $\mathbf{H}_{, 0: d}$ denotes the treatment variable, and $\mathbf{H}_{,(p-1) d: p d}$ denotes the confounders, $\mathbf{w}=\operatorname{softmax}(\mathbf{w})$ to satisfy this constrain Hence, reweighting training samples with the optimal $\mathbf{w}^{*}$ can mitigate the dependence between high-level treatment variable with confounders to the greatest extent. Sui et al.~\cite{sui2022graph} propose a causal attention model that could distinguish between causal and confounding features of a graph. The authors leverage the backdoor adjustment to disentangle the causal and confounding features. The proposed model separates the causal and confounding features from full graphs using attention mechanism.

Bevilacqua et al.~\cite{bevilacqua2021size} leverage causal models inspired by Stochastic Block Models (SBM)~\cite{snijders1997estimation} and graphon random graph models~\cite{lovasz2006limits} to learn size-invariant representations that better extrapolate between test and train graph data. The authors construct graph representations from subgraph densities and use attribute symmetry regularization to mitigate the shift of graph size and vertex attribute distributions. 

Chen et al.~\cite{chen2022invariance} propose to provide guaranteed OOD generalization on graphs under different distribution shifts. The authors leverage three SCMs to characterize the distribution shifts that could happen on graphs. They further argue that GNNs are invariant to distribution shifts if they focus only on a invariant and critical subgraph $G_{c}$ that contains the most of the information in $G$ about the underlying causes of the label. To learn the invariant representation, the authors propose to align with two causal mechanisms that occur during graph generation, i.e., $C \rightarrow G$ and $\left(G_{s}, E_{G}, G_{c}\right) \rightarrow G$ where $C$ represents the causal features, $S$ represents the non-causal features, $E_{G}$ denotes the environment, $G_{c}$ inherits the invariant information of $C$ that would not be affected by the interventions, and $G_{s}$ inherits the varying features. The alignment is realized by decomposing a GNN into a featurizer GNN $g: \mathcal{G} \rightarrow \mathcal{G}_{c}$ aiming to identify the desired $G_{c} ;$ b) a classifier GNN $f_{c}: \mathcal{G}_{c} \rightarrow \mathcal{Y}$ that predicts the label $Y$ based on the estimated $G_{c}$, where $\mathcal{G}_{c}$ refers to the space of subgraphs of $G$. Formally, the learning objectives of $f_{c}$ and $g$ can be formulated as:
$$
\min _{f_{c}, g} R\left(f_{c}\left(\hat{G}_{c}\right)\right) \text {, s.t. } \hat{G}_{c} \perp E, \hat{G}_{c}=g(G),
$$
where $R\left(f_{c}\left(\hat{G}_{c}\right)\right)$ is the empirical risk of $f_{c}$ that takes $\hat{G}_{c}$ as innuts to predict label $Y$ for $G$, and $\hat{G}_{c}$ is the intermediate subgraph containing information about $C$ and is independent of $E$. The final objective is given by,
\begin{equation}
\begin{aligned}
&\max _{f_{c}, g} I\left(\hat{G}_{c} ; Y\right)+I\left(\hat{G}_{s} ; Y\right), \\
&\text { s.t. } \hat{G}_{c} \in \underset{\substack{G_{c}=g(G), \tilde{G}_{c}=g(\tilde{G})}}{\arg \max } I\left(\hat{G}_{c} ; \tilde{G}_{c} \mid Y\right), \\
&I\left(\hat{G}_{s} ; Y\right) \leq I\left(\hat{G}_{c} ; Y\right), \hat{G}_{s}=G-g(G),
\end{aligned}
\end{equation}
This work differs from~\cite{bevilacqua2021size} as it establishes generic SCMs that are compatible with several graph generation models, and different types of distribution shifts.

Wu et al.~\cite{wu2022discovering} aims at identifying graph rationales that capture the invariant causal patterns. To this end, the authors propose  Discovering Invariant Rationales (DIR), which leverages causal interventions to instantiate environments and further distinguish the causal and non-causal parts. The task of invariant rationalization can be formulated as,
\begin{equation}
\min _{h_{\tilde{C}}, h_{\hat{Y}}} \mathcal{R}\left(h_{\hat{Y}} \circ h_{\tilde{C}}(G), Y\right), \quad \text { s.t. } Y \perp \tilde{S} \mid \tilde{C},
\end{equation}
where $h_{\tilde{C}}$ discovers rationale $\tilde{C}$ from the observed $G$, $h_{\hat{Y}}$ represents the classifier, $\tilde{C}$ is the causal rationale and $\tilde{S}=G \backslash \tilde{C}$ is the complement of $\tilde{C}$. Furthermore, to obtain the environments the authors generate $s$-interventional distribution by doing intervention $do(S=s)$ on $S$, which removes every link from the parents $PA(S)$ to the variable $S$ and fixes $S$ to the specific value $s$. By stratifying different values $\mathbb{S}=\{s\}$, they obtain multiple $s$-interventional distributions. The DIR Risk is formulated as, 
\begin{equation}
\begin{aligned}
&\min \mathcal{R}_{\mathrm{DIR}}=\mathbb{E}_{s}[\mathcal{R}(h(G), Y \mid do(S=s))]+\\
&\lambda \operatorname{Var}_{s}(\{\mathcal{R}(h(G), Y \mid do(S=s))\})
\end{aligned}
\end{equation}
where $\mathcal{R}(h(G), Y \mid do(S=s))$ computes the risk under the $s$-interventional distribution, $\operatorname{Var}(\cdot)$ calculates the variance of risks over different $s$-interventional distributions and $\lambda$ is a hyper-parameter to control the strength of invariant learning.

\section{Benchmark, Evaluation, and Code Repositories}
\begin{table*}
\centering
\resizebox{\textwidth}{!}{%
\begin{tabular}{|c|c|c|c|c|c|}
\hline
\textbf{Dataset} & \textbf{Description} & \textbf{Domain} & \textbf{\begin{tabular}[c]{@{}c@{}}Downstream\\ Task\end{tabular}} & \textbf{\begin{tabular}[c]{@{}c@{}}Basic Statistics\\ of the Dataset\end{tabular}} & \textbf{\begin{tabular}[c]{@{}c@{}}Useful in which\\ Category and How?\end{tabular}} \\ \hline
CelebA & \begin{tabular}[c]{@{}c@{}}is a large-scale face attributes dataset. \\ The images in this dataset cover large \\ pose variations and background clutter.\end{tabular} & Vision & \begin{tabular}[c]{@{}c@{}}Face\\ Detection\end{tabular} & \begin{tabular}[c]{@{}c@{}}202,599 celebrity images\\ 10,177 unique celebrities\\ 40 attribute annotations.\end{tabular} & \begin{tabular}[c]{@{}c@{}}Since it provides access to auxiliary labels\\ such as GENDER and SMILE, it has \\ been utilized across multiple causal methods that\\ aim to learn and identify the causal features of an image.\end{tabular} \\ \hline
CK+ & \begin{tabular}[c]{@{}c@{}}The Extended Cohn-Kanade (CK+) dataset \\ contains video sequences ranging from 18 \\ to 50 years of age with various genders\\ and heritage. The videos are labeled to denote\\ the emotions such as anger, disgust, and so on.\end{tabular} & Vision & \begin{tabular}[c]{@{}c@{}}Emotion\\ Detection\end{tabular} & \begin{tabular}[c]{@{}c@{}}593 video sequences\\ 123 different subjects\\ 327 labelled videos\end{tabular} & \begin{tabular}[c]{@{}c@{}}This dataset has been used for causal data \\ augmentation by combining different features \\ useful to predict emotions with \\ confounding features such as noise.\end{tabular} \\ \hline
MNIST & \begin{tabular}[c]{@{}c@{}}The MNIST database (Modified National Institute of \\ Standards and Technology database) \\ is a large collection of handwritten digits.\end{tabular} & Vision & \begin{tabular}[c]{@{}c@{}}Digit\\ Recognition\end{tabular} & \begin{tabular}[c]{@{}c@{}}70,000 instances\\ 10 classes\end{tabular} & \begin{tabular}[c]{@{}c@{}}MNIST is a part of the DigitsDG dataset that\\ contains 4 domains, including MNIST, MNIST-M\\ SVHN and SYN. It has been leveraged by causal\\ methods trying to learn and identify the causal features.\end{tabular} \\ \hline
CMNIST & \begin{tabular}[c]{@{}c@{}}Colored MNIST (CMNIST) is a synthetic\\ binary classification task derived from MNIST.\\ In CMNIST the color and the label have a different\\ correlation in the train set when compared to the\\ correlation in the test set.\end{tabular} & Vision & \begin{tabular}[c]{@{}c@{}}Digit\\ Recognition\end{tabular} & \begin{tabular}[c]{@{}c@{}}70,000 instances\\ 2 classes\end{tabular} & \begin{tabular}[c]{@{}c@{}}CMNIST was a synthetic dataset introduced in the\\ IRM paper. Since the spurrious correlation between\\ color and the label changes in the training and test\\ set, this dataset has been used in a wide-range of\\ works that learn invariance by transfering the\\ causal mechanisms.\end{tabular} \\ \hline
Chest X-Ray14 & \begin{tabular}[c]{@{}c@{}}ChestX-ray14 is a medical imaging dataset\\ which contains frontal-view X-ray images\\ of unique patients along with disease \\ labels mined from text.\end{tabular} & Vision & \begin{tabular}[c]{@{}c@{}}Medical\\ Imaging\end{tabular} & \begin{tabular}[c]{@{}c@{}}112,120 X-ray images\\ 30,805 unique patients \\ 23 years worth data.\end{tabular} & \begin{tabular}[c]{@{}c@{}}Chest X-ray14 is a part of the ChestX-rays dataset\\ that contains 3 domains, including, ChestXray14,\\ MIMIC-CXR, and Stanford CheXpert. These \\ datasets have been leveraged in causal methods \\ trying to learn and identify the causal features.\end{tabular} \\ \hline
Human3.6M & \begin{tabular}[c]{@{}c@{}}Human3.6M is a 3D human pose and corresponding\\ images dataset. It contains various actors both\\ male and female posing in different scenarios.\end{tabular} & Vision & \begin{tabular}[c]{@{}c@{}}Human Pose\\ Estimation\end{tabular} & \begin{tabular}[c]{@{}c@{}}3.6 million images\\ 11 professional actors\\ 17 scenarios\end{tabular} & \begin{tabular}[c]{@{}c@{}}Since this dataset provides access to different poses\\ under different domains, this dataset has been\\ leveraged by causal data augmentation approaches\\ where given the domain and the content, the models generate\\ causally augmented images to train the model.\end{tabular} \\ \hline
CUHK03 & \begin{tabular}[c]{@{}c@{}}The CUHK03 consists of images of different identities,\\ where 6 campus cameras were deployed\\ for image collection and each identity is\\ captured by 2 campus cameras. This dataset \\ provides two types of annotations, one by \\ manually labelled bounding boxes and the other\\ by bounding boxes produced by an automatic detector.\end{tabular} & Vision & \begin{tabular}[c]{@{}c@{}}Person \\ Re-ID\end{tabular} & \begin{tabular}[c]{@{}c@{}}14,097 images\\ 1,467 identities\end{tabular} & \begin{tabular}[c]{@{}c@{}}Since this dataset reflects the same\\ person across multiple backgrounds, this dataset has\\ been leveraged by causal methods that aim to learn\\ and identify the causal features present in an image.\end{tabular} \\ \hline
WaterBirds & \begin{tabular}[c]{@{}c@{}}The WaterBirds dataset consists of water birds\\ and land birds. It was extracted from \\ Caltech-UCSD Birds-200-2011 benchmark dataset,\\ with water and land background \\ extracted from the Places dataset\end{tabular} & Vision & \begin{tabular}[c]{@{}c@{}}Birds \\ Classification\end{tabular} & \begin{tabular}[c]{@{}c@{}}$\sim$5,000 images\\ 2 classes\end{tabular} & \begin{tabular}[c]{@{}c@{}}Since this dataset contains auxiliary labels such \\ as the label for the background of an image,\\ they have been utilized in causal methods \\ that aim to learn and identify the causal \\ features present in an image.\end{tabular} \\ \hline
\begin{tabular}[c]{@{}c@{}}Terra \\ Incognita\end{tabular} & \begin{tabular}[c]{@{}c@{}}This dataset contains images from twenty camera\\ traps which were deployed to monitor animal\\ populations. Since the traps are fixed, the background \\ changes little across images. Capture\\ is triggered automatically, thus eliminating human bias.\end{tabular} & Vision & \begin{tabular}[c]{@{}c@{}}Animals \\ Classification\end{tabular} & \begin{tabular}[c]{@{}c@{}}24,788 images\\ 10 classes\\ 4 domains\end{tabular} & \begin{tabular}[c]{@{}c@{}}Since this dataset contain high cross-domain\\ discrepancies it has been used in causal methods that\\ aim to learn and identify the causal features \\ and by causal methods that capture invariance\\ through the transfer of causal mechanisms\end{tabular} \\ \hline
PACS & \begin{tabular}[c]{@{}c@{}}PACS refers to Photo, Art, Cartoon, and Sketch.\\ Each of the domain have seven categories.\end{tabular} & Vision & \begin{tabular}[c]{@{}c@{}}Object \\ Recognition\end{tabular} & \begin{tabular}[c]{@{}c@{}}9,985 images\\ 4 domains\\ 7 categories\end{tabular} & \begin{tabular}[c]{@{}c@{}}Since this dataset contains the representation\\ of the same object across multiple image styles,\\ it allows models to distinguish between causal \\ and non-causal features. Thus, it is widely used\\ in causal methods that aim to learn and identify\\ the causal features present in an image.\end{tabular} \\ \hline
OfficeHome & \begin{tabular}[c]{@{}c@{}}Office-Home is a benchmark dataset that\\ represents images of the same object under\\ different scenarios, including Art, Clipart,\\ Product and Real-World.\end{tabular} & Vision & \begin{tabular}[c]{@{}c@{}}Object \\ Recognition\end{tabular} & \begin{tabular}[c]{@{}c@{}}15,500 images\\ 65 categories\\ 4 domains\end{tabular} & \begin{tabular}[c]{@{}c@{}}Since this dataset contains the representation\\ of the same object across multiple image styles,\\ it allows models to distinguish between causal\\ and non-causal features. Thus, it is widely used\\ in causal methods that aim to learn and identify\\ the causal features present in an image.\end{tabular} \\ \hline
\end{tabular}%
}
\caption{A description of different vision benchmark datasets and how they have been leveraged by different causality-aware domain generalization methods.}
\label{datasets}
\end{table*}

\begin{table*}
\centering
\resizebox{\textwidth}{!}{%
\begin{tabular}{|c|c|c|c|c|c|}
\hline
\textbf{Dataset} & \textbf{Description} & \textbf{Domain} & \textbf{\begin{tabular}[c]{@{}c@{}}Downstream\\ Task\end{tabular}} & \textbf{\begin{tabular}[c]{@{}c@{}}Basic Statistics\\ of the Dataset\end{tabular}} & \textbf{\begin{tabular}[c]{@{}c@{}}Useful in which\\ Category and How?\end{tabular}} \\ \hline
PROTEINS & \begin{tabular}[c]{@{}c@{}}PROTEINS is a dataset of proteins that are\\ classified as enzymes or non-enzymes.\\ Nodes represent the amino acids and two nodes \\ are connected by an edge if they are\\ less than 6 Angstroms apart.\end{tabular} & Graph & \begin{tabular}[c]{@{}c@{}}Graph \\ Classification\end{tabular} & \begin{tabular}[c]{@{}c@{}}1,113 graphs\\ 39 Avg. no. of nodes\\ 2 classes\end{tabular} & \begin{tabular}[c]{@{}c@{}}This dataset is used along with other biological\\ graph datasets such as MUTAG and NCI1 \\ by causal methods that aim to learn\\ and identify the causal features present in an image.\end{tabular} \\ \hline
Multi-NLI & \begin{tabular}[c]{@{}c@{}}The Multi-Genre Natural Language Inference\\ dataset is a collection of written and spoken\\ english data over ten different genres.\end{tabular} & Text & \begin{tabular}[c]{@{}c@{}}Natural \\ Language \\ Inference\end{tabular} & \begin{tabular}[c]{@{}c@{}}433K sentence-pairs\\ 10 domains\end{tabular} & \begin{tabular}[c]{@{}c@{}}This dataset contains spurious correlations as usually, \\ the second sentence in a pair containing a negation\\ is classified as contradiction. Works leveraging\\ causal data augmentation have tried to tackle this spuriousness\\ by generating counterfactuals for this situation.\end{tabular} \\ \hline
CivilComments & \begin{tabular}[c]{@{}c@{}}Civil Comments contains the archive\\ of the Civil Comments platform. It is annotated\\ for toxicity across different demographic\\ groups in the english language.\end{tabular} & Text & \begin{tabular}[c]{@{}c@{}}Toxicity \\ Detection\end{tabular} & \begin{tabular}[c]{@{}c@{}}$\sim$2 million comments\\ 8 domains (demographic\\ groups)\\ 2 classes\end{tabular} & \begin{tabular}[c]{@{}c@{}}This dataset has been leveraged by causal methods\\ that aim to learn and identify the causal features\\ from text as it is possible to find the spurious\\ correlations between different words and the\\ toxicity labels of the comments.\end{tabular} \\ \hline
\begin{tabular}[c]{@{}c@{}}Amazon Kindle\\ Reviews\end{tabular} & \begin{tabular}[c]{@{}c@{}}The Amazon Kindle Reviews dataset\\ contains product reviews from\\ kindle. The main goal is to infer the\\ review's sentiment as positive or negative.\end{tabular} & Text & \begin{tabular}[c]{@{}c@{}}Sentiment \\ Classification\end{tabular} & 10,500 reviews & \begin{tabular}[c]{@{}c@{}}This dataset is leveraged by the causal data\\ augmentation methods as these models perform data\\ augmentations on non-causal (non-sentimental)\\ words to study the domain generalization problem.\\ It has also been used by causal methods that\\ aim to learn and identify the causal features\\ from text as the sentimental words have a causal\\ relation to the sentiment classification task.\end{tabular} \\ \hline
\end{tabular}%
}
\caption{A description of different Graph and Text benchmark datasets and how they have been leveraged by different causality-aware domain generalization methods.}
\label{GT_datasets}
\end{table*}

In this section, we provide a comprehensive review of benchmark datasets and evaluation metrics for all three types of data, i.e., image, text and graph, for the causal domain generalization task.
\subsection{Benchmark Datasets}
Causality-aware domain generalization has been studied across various applications, including but not limited to computer vision, natural language processing, and graphs. Tables~\ref{datasets} and~\ref{GT_datasets} summarize the commonly used datasets based on the different applications. In this section, we briefly describe these datasets and the applications.

\noindent{\textbf{Face Detection}} can be decomposed into multiple tasks, such as face attributes detection, and human face synthesis. Some of the benchmark datasets are CelebA~\cite{karras2017progressive} which contains auxiliary attribute labels (such as $GENDER$, $SMILE$) to improve the generalization performance. A range of datasets, including Face-Forensics++~\cite{rossler2019faceforensics} and DeeperForensics-1.0~\cite{jiang2020deeperforensics}, were leveraged to facilitate generalization for the human face synthesis task. 

\noindent{\textbf{Emotion detection}} has been studied extensively to aid applications such as mental health care and driver drowsiness detection. Benchmark datasets such as CK+~\cite{lucey2010extended}, MMI~\cite{valstar2010induced}, and Oulu-CASIA~\cite{zhao2011facial} have been leveraged to test generalization capabilities for emotion detection. These datasets contain multiple face angles along with basic expression labels such as $anger$, $disgust$, and $fear$.

\noindent{\textbf{Handwritten digit recognition}} are a good example of distribution shifts as the different writing styles of people are as different domains and the shape of the digit remains invariant. Many methods have leveraged benchmark digits datasets to solve the generalization problem in digits recognition. Some majorly applied datasets are FashionMNIST~\cite{xiao2017fashion} (which is a collection of grayscale fashion article images), ColoredMNIST+~\cite{guo2021out}, MNIST-M~\cite{ganin2015unsupervised}, SVHN~\cite{netzer2011reading}, and SYN~\cite{ganin2015unsupervised}, MNIST~\cite{lecun1998gradient} and its variants such as ColoredMNIST~\cite{arjovsky2019invariant} (where the digits have different color distributions) and RotatedMNIST~\cite{ghifary2015domain} (where the digits are rotated on different angles).

\noindent{\textbf{Medical Imaging}} refers to an umbrella term comprising multiple tasks. In the medical scenario, domain shifts are mainly caused by different acquisition processes. Thus, to improve the generalization performance, a range of works~\cite{mahajan2021domain, sun2020latent, ouyang2021causality} leveraged real-world medical imaging datasets such as Alzheimer's Disease Neuroimaging Initiative (ADNI)~\cite{jack2008alzheimer}. Furthermore, the Chest X-rays dataset contains images from three sources: NIH~\cite{wang2017chestx}, ChexPert~\cite{irvin2019chexpert} and RSNA~\cite{franquet2018imaging}. The central task for this dataset was to classify the image to whether the patient has Pneumonia (1) or not (0).

\noindent{\textbf{Body pose estimation}} refers to the 3D pose estimation task. Recently the authors et al.~\cite{zhang2021learning1} proposed to leverage causality to improve the cross-domain pose estimation problem. They utilize multiple benchmark datasets such as Human3.6M~\cite{ionescu2013human3}, 3DPW~\cite{von2018recovering}, MPI-INF-3DHP (3DHP)~\cite{mehta2017monocular}, SURREAL~\cite{varol2017learning}, and HumanEva~\cite{sigal2010humaneva} for body pose estimation.

\noindent{\textbf{Person Re-IDentification}} aims at matching person images of the same identity across multiple camera views. In this scenario, the domain shift arises in image resolution, viewpoint, lighting condition, background, etc. Some benchmark datasets for this task are CUHK02~\cite{li2013locally}, CUHK03~\cite{li2014deepreid}, Market1501~\cite{zheng2015scalable}, DukeMTMC-ReID~\cite{zheng2017unlabeled}, CUHKSYSU PersonSearch~\cite{xiao2016end}. 

\noindent{\textbf{Animal and Birds Classification}} refers to classifying different animals and bird species observed across multiple environments. For instance, the WaterBirds dataset contains images of water birds (Gulls) and land birds (Warblers) extracted from the Caltech-UCSD Birds-200- 2011 (CUB) dataset~\cite{wah2011caltech} with water and land background extracted from the Places dataset~\cite{zhou2017places}. In addition, there are multiple datasets for animal classification, such as iWiLDSCam~\cite{beery2019iwildcam}, and  TerraIncognita~\cite{beery2018recognition}. 

\noindent{\textbf{Object Recognition}} is one of the most prominent tasks for studying the domain generalization problem, where the domain shift varies substantially across different datasets. For instance, PACS~\cite{li2017deeper}, OfficeHome~\cite{venkateswara2017deep}, DomainNet~\cite{peng2019moment} and ImageNet-R~\cite{hendrycks2021many}, deal with image style changes where the same object is varied across different styles. Another commonly used datasets are DomainNet~\cite{peng2019moment}, VLCS~\cite{fang2013unbiased}, ImageNet-C~\cite{hendrycks2019benchmarking}, NICO~\cite{he2021towards} and NICO++~\cite{zhang2022nico++}.

\noindent{\textbf{Graph Classification}} is a crucial activity when dealing with graph distribution shifts. For this endeavor, a variety of datasets have been chosen to help with domain generalization. One such small-scale real-world dataset is HIV , which was modified from MoleculeNet~\cite{wu2018moleculenet}. ZINC is a real-world dataset for molecular property regression from the ZINC database~\cite{gomez2018automatic}. Motif is a synthetic base-motif dataset motivated by Spurious-Motif~\cite{wu2022discovering}.

\noindent{\textbf{Node Classifciation}} is another prominent task when dealing with node classification. Currently there exists several datasets that contain the out-of-distribution samples to test graph generalization capabilities, such as Cora~\cite{bojchevski2017deep}, and Arxiv~\cite{hu2020open}. There also exist synthetic dataset such CBAS derived from the BA-Shapes~\cite{ying2019gnnexplainer}.

\noindent{\textbf{Sentiment Classification}} deals with the task of classifying sentiments from human documents. Currently there exists several benchmark datasets for this task, examples include, Amazon Reviews Dataset~\cite{fang2014domain}, IMDb dataset~\cite{maas2011learning}, FineFood dataset~\cite{mcauley2013amateurs}. Various works such as~\cite{choi2022c2l} aim to utilize multiple datasets listed here by training the model on one of the benchmark datasets and evaluate it against the other benchmarks.

\noindent{\textbf{Toxicity Detection}} refers to the task of detecting toxicity in textual data. One of the well-known benchmarks is the CivilComments dataset~\cite{borkan2019nuanced}. When evaluating for the cross-domain generalizability, the model is tested to see whether the model can detect toxicity without depending on the demographic identities.

\noindent{\textbf{Natural Language Inference}} aims at classifying pair of texts based on their logical relationship. One of the benchmark datasets for NLI is MultiNLI~\cite{williams2017broad}. A series of works such as ~\cite{choi2022c2l} have utilized this benchmark.

\subsection{Evaluation}
Causality-aware domain generalization methods employ evaluation mechanisms that are very similar to standard domain generalization methods. Furthermore, the causal theories and methodologies employed by these frameworks help in the identification of invariant relationships while doing the same downstream task. Thus, the evaluation procedure is determined by the technique's nature, i.e., whether the approach is a single source domain generalization (where the model is trained on data from a single source) or a multi-source domain generalization (where the model is trained on data from multiple sources).

For instance, the \textit{leave-one-domain-out rule} is one of the most prominent when dealing with multi-source domain generalization~\cite{yang2021causalvae,mahajan2021domain,heinze2017conditional,lu2021invariant}. Per this rule, given a dataset containing at least two distinct domains, multiple are used as source domains for training the model while one is used as the target domain, on which the model is directly tested without any adaptation. Another commonly employed strategy for domain generalization is \textit{Training-domain validation set}~\cite{wang2021improving,ouyang2021causality}. In this setting, the source domains are split into two parts; one is used for training while the other is used for validation. While training the model, each domain's training parts are combined, and the validation parts are used to select the best model. When dealing with single domain generalization methods, the popular approach is to train the model on the source domain and directly evaluate it on the different test domains.

Classification accuracy, top-1 error rate, top-5 error rate, and DICE scores are some of the main measures used to assess these models. The classification accuracy metric indicates how well the model classifies samples. The top-1 error rate determines if the model's projected top class corresponds to the target label. In the instance of the Top-5 mistake rate, we examine whether or not the target label appears in the top-5 predictions. Finally, the DICE scores compare the pixel-by-pixel agreement of a projected segmentation with its matching ground truth. Aside from these commonly used metrics, some works~\cite{mitrovic2020representation} leveraged mean Corruption Error (mCE) and mean relative Corruption Error (mrCE), which are commonly used to evaluate a model's robustness. 

Although the causality-aware models strive to perform the same downstream job, we believe that causal evaluation procedures and metrics, in addition to the traditional setting, are required to verify the models' causal features. For example, Yang et.al~\cite{yang2021causalvae} leveraged  Maximal Information Coefficient (MIC) and Total Information Coefficient (TIC)~\cite{kinney2014equitability} as additional evaluation metrics. They use these metrics because they reflect the degree of information relevance between the learnt representation and the ground truth labels of ideas, which is important because their model tries to learn causal representations.
\section{Conclusion}
In this survey, we provide a comprehensive overview of out-of-distribution generalization approaches from a causal perspective. In particular, depending on at which stage of training the machine learning framework the causal domain generalization component is applied, we classify them into three main categories, namely causal data augmentation methods (applied during the data pre-processing phase), invariant causal representation learning approaches (performed during the representation learning stage), and invariant causal mechanism learning algorithms (applied at the classifier level) and explain the state-of-the-art methods in each category. Depending on the approach taken, we further categorize the approaches in each category into sub-categories as shown in
Figure~\ref{categorization}. Moreover, we extend our categories to the textual and graph data and classify the approaches developed for those data types into our three categories. Comparing the comprehensive body of literature for the image data with recent works on textual and graph data, we observe many future research directions for these data types. Specifically, while most works on these data types belong to the causal data augmentation category, the causal representation learning and causal invariant mechanism learning-based approaches are greatly underexplored. We therefore suggest exploring these two directions for both of these data types.
To evaluate the causal domain generalization approaches systematically, we provide a comprehensive list of commonly-used datasets and evaluation metrics to assess the performance of the proposed frameworks. These evaluation guidelines can be used by researchers and practitioners to appropriately evaluate the performance of their frameworks and compare the performance of existing methodologies.

\bibliographystyle{unsrt}
\bibliography{template}  





\end{document}